\documentclass{article}

\usepackage{caption}
\usepackage{graphicx}
\usepackage{tcolorbox}
\usepackage[final]{mystyle}

\usepackage[utf8]{inputenc} 
\usepackage[T1]{fontenc}    
\usepackage{url}            
\usepackage{booktabs}       
\usepackage{amsfonts}       
\usepackage{nicefrac}       
\usepackage{microtype}      
\usepackage{xcolor}         

\usepackage{amsmath}
\usepackage{amssymb}
\usepackage{subfig}
\usepackage{diagbox}
\usepackage{epsfig}
\usepackage{extpfeil}
\usepackage{wrapfig}
\usepackage{multirow}
\usepackage{wrapfig}
\usepackage{color,xcolor,colortbl}
\usepackage{pifont}
\usepackage{xspace}
\usepackage{float}
\usepackage[pagebackref=false,breaklinks=true,colorlinks=true,bookmarks=false,citecolor=blue,linkcolor=red]{hyperref}
\definecolor{lightgray}{gray}{0.95}
\definecolor{color3}{gray}{0.95}
\definecolor{rouse}{rgb}{0.981,0.961,0.941}
\definecolor{light-yellow}{rgb}{1,1,0.93}
\definecolor{light-green}{rgb}{0.95,1,0.95}

\definecolor{color3}{rgb}{0.95,0.95,0.95}
\newcommand{\cmark}{\ding{51}}%
\newcommand{\xmark}{\ding{55}}%

\newcommand{\dataname}{\emph{Motion-X}\xspace}

\title{Motion-X: A Large-scale 3D Expressive \\Whole-body Human Motion Dataset}

\author{
    Jing Lin$^{1,2}$\thanks{Equal Contribution}~~, 
    Ailing Zeng$^{1*}$\thanks{Corresponding author}~~, 
    Shunlin Lu$^{1,3*}$, \\
    \textbf{Yuanhao Cai$^{2}$}, 
    \textbf{Ruimao Zhang$^{3}$}, 
    \textbf{Haoqian Wang$^{2}$},
    \textbf{Lei Zhang$^{1}$} \\
    \textsuperscript{1}International Digital Economy Academy (IDEA) \quad \\
    \textsuperscript{2}Shenzhen International Graduate School, Tsinghua University \quad \\
    \textsuperscript{3}The Chinese University of Hong Kong, Shenzhen \quad \\
    \url{https://github.com/IDEA-Research/Motion-X}
}

\begin{document}

\maketitle

\vspace{-0.6cm}
\begin{abstract}
\vspace{-0.2cm}
In this paper, we present Motion-X, a large-scale 3D expressive whole-body motion dataset. 
Existing motion datasets predominantly contain body-only poses, lacking facial expressions, hand gestures, and fine-grained pose descriptions. 
Moreover, they are primarily collected from limited laboratory scenes with textual descriptions manually labeled, which greatly limits their scalability.
To overcome these limitations, we develop a whole-body motion and text annotation pipeline, which can automatically annotate motion from either single- or multi-view videos 
and provide comprehensive semantic labels for each video and fine-grained whole-body pose descriptions for each frame. 
This pipeline is of high precision, cost-effective, and scalable for further research. 
Based on it, we construct Motion-X, which comprises 15.6M precise 3D whole-body pose annotations (i.e., SMPL-X) covering 81.1K motion sequences from massive scenes. 
Besides, Motion-X provides 15.6M frame-level whole-body pose descriptions and 81.1K sequence-level semantic labels.
Comprehensive experiments demonstrate the accuracy of the annotation pipeline and the significant benefit of Motion-X in enhancing expressive, diverse, and natural motion generation, as well as 3D whole-body human mesh recovery. 

\vspace{-0.2cm}
\end{abstract}

\section{Introduction}
\vspace{-0.1cm}

Human motion generation aims to automatically synthesize natural human movements. It has wide applications in robotics, animation, games, and generative creation. Given a text description or audio command, motion generation can be controllable to obtain the desired human motion sequence. Text-conditioned motion generation has garnered increasing attention in recent years since it behaves in a more natural interactive way~\cite{ahuja2019language2pose,mld,posescript,humanml3d,temos,kit,plappert2018learning,babel,motiondiffuse,modiff}. 

\begin{figure*}[ht]
	\begin{center}
		\begin{tabular}[t]{c} \hspace{-4.6mm}
			\includegraphics[width=1\textwidth]{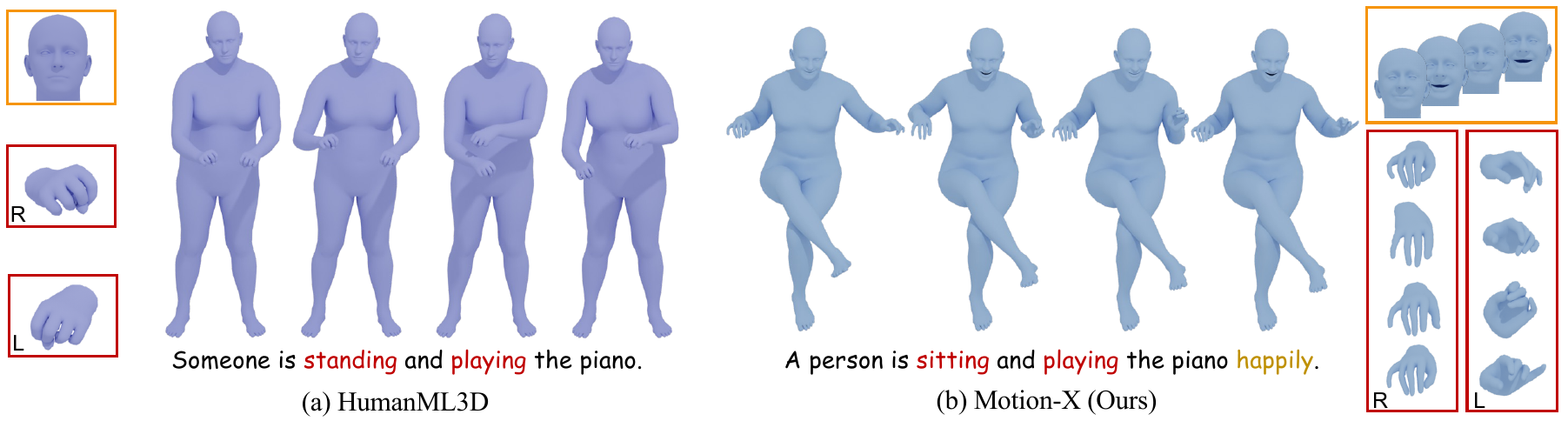}
		\end{tabular}
	\end{center}
	\vspace*{-0.3cm}
	\caption{\small Different from (a) previous motion dataset~\cite{humanml3d,babel}, 
    (b) our dataset captures body, facial expressions, and hand gestures. 
    We highlight the comparisons of facial expressions and hand gestures. 
    }
  \label{fig:teaser}
  \vspace{-0.3cm}
\end{figure*}

Although existing text-motion datasets~\cite{humanml3d,amass,kit,babel} have greatly facilitated the development of motion generation~\cite{mld,Ho2020DenoisingDP,Song2020DenoisingDI,Tevet2022HumanMD,motiondiffuse}, their scale, diversity, and expressive capability remain unsatisfactory.
Imagine generating ``\emph{a man is playing the piano happily}", as depicted in Fig.~\ref{fig:teaser}(a), the motion from existing dataset~\cite{humanml3d} only includes the body movements, without finger movements or facial expressions.
The missing hand gestures and facial expressions severely hinder the high level of expressiveness and realism of the motion. Additionally, certain specialized motions, such as high-level skiing, aerial work, and riding are challenging to be captured in indoor scenes.
To sum up, existing datasets suffer from four main limitations: 1) body-only motions without facial expressions and hand poses; 2) insufficient diversity and quantity, only covering indoor scenes; 3) lacking diverse and long-term motion sequences; and 4) manual text labels that are unscalable, unprofessional and labor-intensive. These limitations hinder existing generation methods to synthesize expressive whole-body motion with diverse action types. Therefore, \emph{how to collect large-scale whole-body motion and text annotations from multi-scenarios are critical in addressing the data scarcity issue.}

Compared to indoor marker-based mocap systems, markerless vision-based motion capture methods~\cite{osx,li2021aist,GyeongsikMoon2020hand4whole,YufeiXu2022ViTPoseSV,glamr,yang2023edpose} become promising to capture large-scale motions from massive videos. Meanwhile, human motion can be regarded as a sequence of kinematic structures, which can be automatically translated into pose scripts using rule-based techniques~\cite{posescript}. More importantly, although markerless capture (e.g., pseudo labels) is not as precise as marker-based methods, collecting massive and informative motions, especially local motions, could still be beneficial~\cite{pang2022benchmarking,osx,moon2022neuralannot,moon2023three,yi2023generating}. Besides, text-driven motion generation task requires semantically corresponding motion labels instead of vertex-corresponding mesh labels, and thus have a higher tolerance of motion capture error.
Bearing these considerations in mind, we design a scalable and systematic pipeline for motion and text annotation in both multi-view and single-view videos. 
Firstly, we gather and filter massive video recordings from a variety of scenes with challenging, high-quality, multi-style motions and sequence-level semantic labels.
Subsequently, we estimate and optimize the parameters of the SMPL-X model~\cite{smpl-x} for the whole-body motion annotation. Due to the depth ambiguity and various challenges in different scenes, existing monocular estimation models typically fail to yield satisfactory results. To address this issue, we systematically design a high-performance framework incorporating several innovative techniques, including a hierarchical approach for whole-body keypoint estimation, a score-guided adaptive temporal smoothing and optimization scheme, and a learning-based 3D human model fitting process. By integrating these techniques, we can accurately and efficiently capture the ultimate 3D motions.
Finally, we design an automatic algorithm to caption frame-level descriptions of whole-body poses. We obtain the body and hand scripts by calculating spatial relations among body parts and hand fingers based on the SMPL-X parameters and extract the facial expressions with an emotion classifier. We then aggregate the low-level pose information and translate it into textual pose descriptions.

Based on the pipeline, we collect a large-scale whole-body expressive motion dataset named \dataname, which includes 15.6M frames and 81.1K sequences with precise 3D whole-body motion annotations, pose descriptions, and semantic labels. To compile this dataset, we collect massive videos from the Internet, with a particular focus on game and animation motions, professional performance, and diverse outdoor actions. Additionally, we incorporated data from eight existing action datasets~\cite{cai2022humman,chung2021haa500,liu2019ntu,amass,taheri2020grab,tsuchida2019aist,baum,zhang2022egobody}.
Using \dataname, we build a benchmark for evaluating several state-of-the-art (SOTA) motion generation methods. Comprehensive experiments demonstrate the benefits of \dataname for diverse, expressive, and realistic motion generation (shown in Fig.~\ref{fig:teaser} (b)). Furthermore, we validate the versatility and quality of \dataname on the whole-body mesh recovery task.

Our contributions can be summarized as follows:
\begin{itemize}
\vspace{-0.1cm}
\item We propose a large-scale expressive motion dataset with precise 3D whole-body motions and corresponding sequence-level and frame-level text descriptions.
\item We elaborately design a automatic motion and text annotation pipeline, enabling efficient capture of high-quality human text-motion data at scale.
\item Comprehensive experiments demonstrate the accuracy of the motion annotation pipeline and the benefits of \dataname in 3D whole-body motion generation and mesh recovery tasks.

\end{itemize}

\section{Preliminary and Related Work}

In this section, we focus on introducing existing \textbf{datasets} for human motion generation. For more details about the motion generation methods, please refer to the appendix.

Benchmarks annotated with sequential human motion and text are mainly collected for three tasks: action recognition~\cite{carreira2019short,chung2021haa500,gu2018ava,liu2019ntu,shahroudy2016ntu,trivedi2021ntu}, human object interaction~\cite{hassan2021stochastic,hassan2019resolving,li2019hake,taheri2020grab,zhang2022egobody,zheng2022gimo}, and motion generation~\cite{humanml3d,guo2020action2motion,amass,kit,babel,yi2023generating}.
Specifically, KIT Motion-Language Dataset~\cite{kit} is the first public dataset with human motion and language descriptions, enabling multi-modality motion generation~\cite{ahuja2019language2pose,temos}. Although several indoor human motion capture (mocap) datasets have been developed~\cite{Gross2001TheCM,Ionescu_2014_hm36,Sigal2010HumanEvaSV,Trumble2017TotalC3}, they are scattered. AMASS~\cite{amass} is noteworthy as it collects and unifies 15 different optical marker-based mocap datasets to build a large-scale motion dataset through a common framework and parameterization via SMPL~\cite{smpl}. This great milestone benefits motion modeling and its downstream tasks. Additionally, BABEL~\cite{babel} and HumanML3D~\cite{humanml3d} contribute to the language labels through crowdsourced data collection. BABEL proposes either sequence labels or sub-sequence labels for a sequential motion, while HumanML3D collects three text descriptions for each motion clip from different workers.
Thanks to these text-motion datasets, various motion generation methods have rapidly developed and shown advantages in diverse, realistic, and fine-grained motion generation~\cite{mld,Tevet2022HumanMD,Yuan2022PhysDiffPH,t2m-gpt,motiondiffuse,modiff}. 

However, existing text-motion datasets have several limitations, including the absence of facial expressions and hand gestures, insufficient data quantity, limited diversity of motions and scenes, coarse-grained and ambiguous descriptions, and the lack of long sequence motions. To bridge these gaps, we develop a large-scale whole-body expressive motion dataset with comprehensive sequence- and frame-level text labels. We aim to address these limitations and open up new possibilities for future research. We provide quantitative comparisons of \dataname and existing datasets in Tab.~\ref{tab:comp}.

\begin{table*}[t]
	\centering
	\resizebox{1\linewidth}{!}
	{%
		\begin{tabular}{l|cccc|ccc|ccc}
			\toprule
      \rowcolor{lightgray}
			 & \multicolumn{4}{c|}{\text{Motion Annotation}} &  \multicolumn{3}{c|}{\text{Text Annotation}}  &  \multicolumn{3}{c}{\text{Scene}}\\ 
			\rowcolor{lightgray}
      \multirow{-2}{*}{\text{Dataset}}
      ~ & ~~~Clip~~~ & ~~~Hour~~~ & Whole-body?  &{Source} & ~~~Motion~~~ & ~~~Pose~~~ & Whole-body?&Indoor &Outdoor&RGB \\ 
			\midrule
			KIT-ML'16~\cite{kit} & 3911 & 11.2 & \xmark &{Marker-based MoCap} & 6278 & 0 & \xmark & \cmark &\xmark&\xmark \\ 
			AMASS'19~\cite{amass} & 11265 & 40.0 & \xmark &{Marker-based MoCap} & 0 & 0 & \xmark & \cmark &\xmark&\xmark \\ 
			BABEL'21~\cite{babel} & 13220 & 43.5 & \xmark &{Marker-based MoCap} &91408 & 0 & \xmark & \cmark &\xmark&\xmark \\
                Posescript'22~\cite{posescript}&-&-&\xmark&{Marker-based MoCap} &0& 120k&\xmark&\cmark &\xmark&\xmark\\
			HumanML3D'22~\cite{humanml3d} & 14616 & 28.6 & \xmark &{Marker-based MoCap} & 44970 & 0 & \xmark & \cmark &\xmark&\xmark \\ \midrule
			\bf Motion-X (Ours) & 81084 & 144.2 & \cmark &{Pseudo GT \& MoCap} & 81084 & 15.6M & \cmark & \cmark & \cmark&\cmark \\ \bottomrule
	\end{tabular}}
        \vspace{0.05cm}
	\caption{\small Comparisons between \dataname and existing text-motion datasets. The first column shows the name and public year of datasets. \dataname provides both indoor and outdoor whole-body motion and text annotations.}
	\label{tab:comp}
 \vspace{-0.5cm}
\end{table*}

\section{Motion-X Dataset}

\subsection{Overview}

\begin{figure}[h]
    \begin{minipage}{0.53\textwidth}
    \centering
    \resizebox{\linewidth}{!}{
        \makeatletter\def\@captype{table}\makeatother
        \vspace{-0.2cm}
        \begin{tabular}{lccccc}
            \toprule
                \rowcolor{color3} \text{Data} &~~\text{Clip}~~ & ~\text{Frame}~ & {\text{GT Motion}} & {\text{P-GT Motion}} & \text{Text} \\ \midrule
                AMASS~\cite{humanml3d} & 26.3K & 5.4M & B &{H, F} & S \\ 
                HAA500~\cite{chung2021haa500} & 5.2K & 0.3M & - &{B, H, F}& S \\ 
                AIST~\cite{tsuchida2019aist} & 1.4K & 0.3M & - &{B, H, F} & S \\
                HuMMan~\cite{cai2022humman} & 0.7K & 0.1M & - &{B, H, F} & S \\
                GRAB~\cite{taheri2020grab} & 1.3K & 0.4M & B,H &{F} & S \\ 
                EgoBody~\cite{zhang2022egobody} & 1.0K & 0.4M & B,H &{F} & - \\ 
                BAUM~\cite{baum} & 1.4K & 0.2M & - &{F} & S \\ 
                IDEA400* & 12.5K & 2.6M & - &{B, H, F} & - \\
                Online Videos* & 32.5K & 6.0M & - &{B, H, F} & -\\ \midrule
                Motion-X & 81.1K & 15.6M & B, H &{B,H,F} & S,P  \\ \bottomrule
            \end{tabular}
        }
        \vspace{-0.1cm}
        \captionof{table}{\small Statistics of sub-datasets. B, H, F are body, hand, and face. S and P are semantic and pose texts. P-GT is pseudo ground truth. * denotes videos are collected by us.}
        \vspace{-0.2cm}
        \label{tab:stat}
    \end{minipage}
    \hfill
    \begin{minipage}{0.42\textwidth}
    \centering
    \vspace{-0.1cm}
    \includegraphics[width=6.2cm]{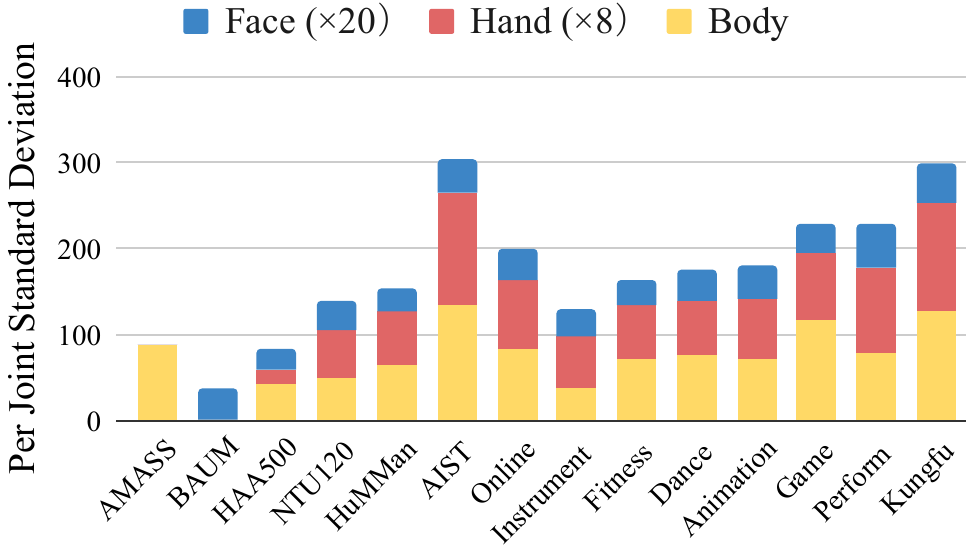}
    \caption{\small Diversity statistics of the face, hand, and body motions in each subdatasets.}
    \label{fig:data_comp} 
    \end{minipage}
\end{figure}

As shown in Tab.~\ref{tab:stat}, we collect \dataname from eight datasets and online videos and provide the following motion and text annotations:
15.6M 3D whole-body SMPL-X annotation,
81.1K sequence-level semantic descriptions (e.g., walking with waving hand and laughing), 
and frame-level whole-body pose descriptions. Notably, original sub-datasets lack either whole-body motion or text labels and we unify them with our annotation pipeline. 
All annotations are manually checked to guarantee quality. In Fig.~\ref{fig:data_comp}, we show the averaged temporal standard deviation of body, hand, and face keypoints of each sub-dataset, highlighting the diversity of hand movements and facial expressions, which fills in the gaps of previous body-only motion data.

\begin{figure*}[t]
\begin{center}
    \includegraphics[width=1\textwidth]{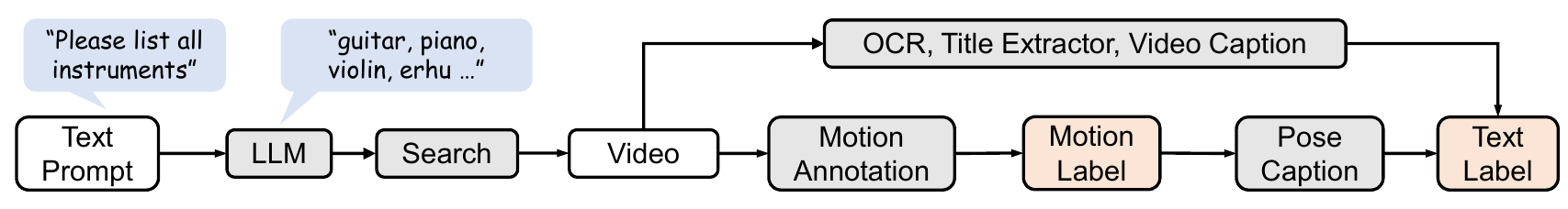}
\end{center}
\vspace{-0.3cm}
\caption{Illustration of the overall data collection and annotation pipeline.}
\label{figure:overall_data_collection}
\vspace{-0.4cm}
\end{figure*}

\begin{figure*}[h]
  \begin{center}
      \includegraphics[width=0.99\textwidth]{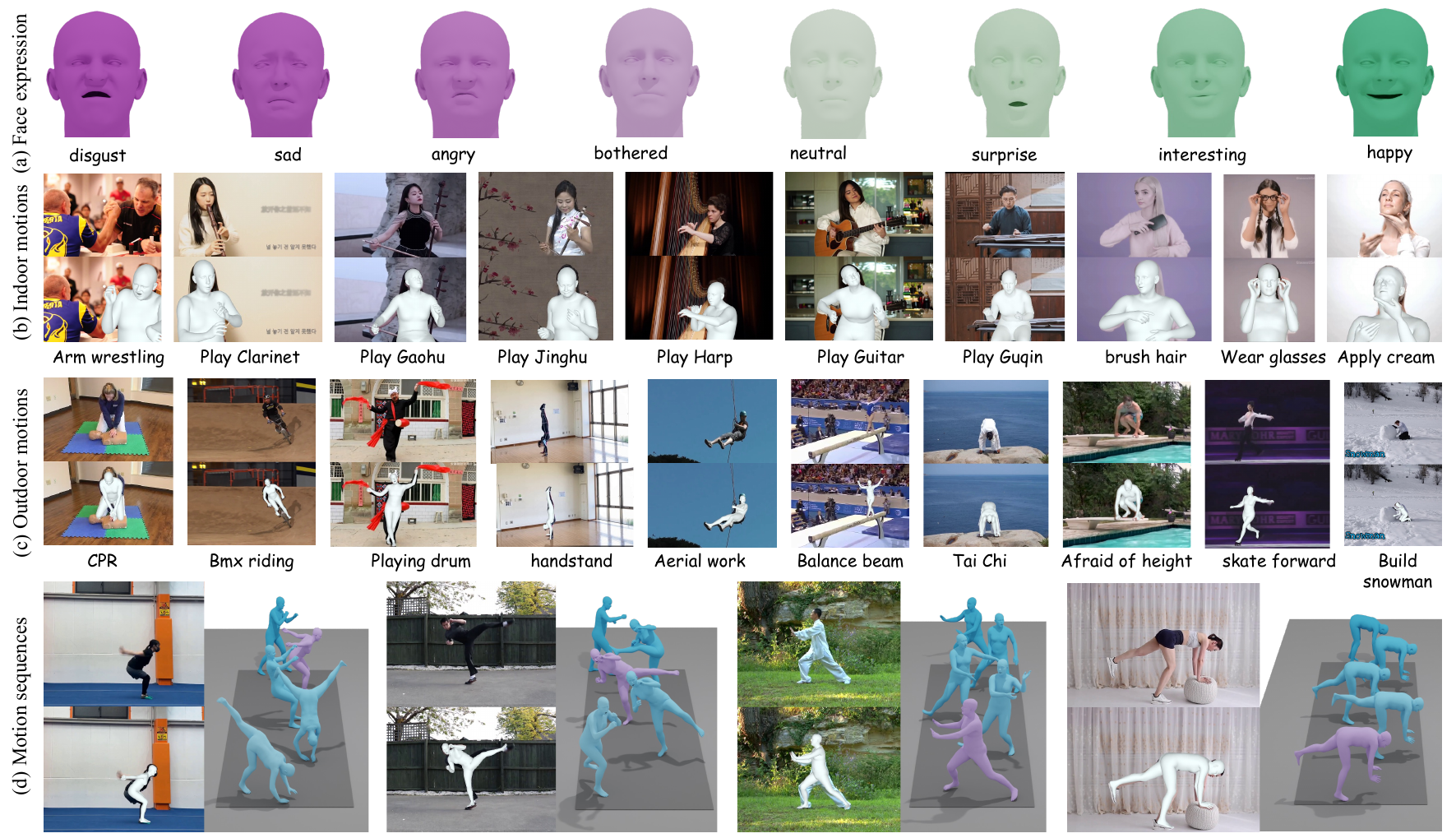}
  \end{center}
  \vspace{-0.3cm}
  \caption{\small Overview of \dataname. It includes: (a) diverse facial expressions extracted from BAUM~\cite{baum}, (b) indoor motion with expressive face and hand motions, (c) outdoor motion with diverse and challenging poses, and (d) several motion sequences. Purple SMPL-X is the observed frame, and the others are neighboring poses.}
  \vspace{-0.6cm}
  \label{fig:dataset}
  \end{figure*}

\subsection{Data Collection}
As illustrated in Fig.~\ref{figure:overall_data_collection}, the overall data collection pipeline involves six key steps: 1) designing and sourcing motion text prompts via large language model (LLM)~\cite{jiao2023chatgpt}, 2) collecting videos, 3) preprocessing candidate videos through human detection and video transition detection, 4) capturing whole-body motion (Sec.~\ref{sec:lable_motion}), 5) captioning sequence-level semantic label and frame-level whole-body pose description(Sec.~\ref{sec:lable_text}), and 6) performing the manual inspection. 

We gather 37K motion sequences from existing datasets using our proposed unified annotation framework, including the multi-view datasets (AIST~\cite{tsuchida2019aist}), human-scene-interaction datasets (EgoBody~\cite{zhang2022egobody} and GRAB~\cite{taheri2020grab}), single-view action recognition datasets (HAA500~\cite{chung2021haa500}, HuMMan~\cite{cai2022humman}), and body-only motion capture dataset (AMASS~\cite{amass}). 
For these datasets, steps 1 and 2 are skipped. Notably, only EgoBody and GRAB datasets provide SMPL-X labels with body and hand pose, thus we annotate the SMPL-X label for the other motions. For AMASS, which contains the body and roughly static hand motions, we skip step 4 and fill in the facial expression with a data augmentation mechanism. The facial expressions are collected from a facial datasets BAUM~\cite{baum} via a face capture and animation model EMOCA~\cite{emoca}.
To enrich the expressive whole-body motions, we record an dataset IDEA400, which provides 13K motion sequences covering 400 diverse actions. Details about the processing of each sub-dataset and IDEA400 are available in the appendix. 

To improve the appearance and motion diversity, we collect 32.5K monocular videos from online sources, covering various real-life scenes as depicted in Fig.~\ref{fig:dataset}.
Since human motions and actions are context-dependent and vary with the scenario,
%
we design action categories as motion prompts based on the scenario and function of the action via LLM. To ensure comprehensive coverage of human actions, our dataset includes both general and domain-specific scenes. The general scenes encompass daily actions (e.g., brushing hair, wearing glasses, and applying creams), sports activities (e.g., high knee, kick legs, push-ups), various musical instrument playing, and outdoor scenes (e.g., BMX riding, CPR, building snowman). The inclusion of general scenes helps bridge the gap between existing data and real-life scenarios.
In addition, we incorporate domain-specific scenes that require high professional skills, such as dance, Kung Fu, Tai Chi, performing arts, Olympic events, entertainment shows, games, and animation motions. Based on the prompts describing the above scenes, we run the collection pipeline to gather the data from online sources for our dataset.

\section{Automatic Annotation Pipeline}
\label{sec:auto}
\subsection{Universal Whole-body Motion Annotation}
\label{sec:lable_motion}

\begin{figure*}[t]
\begin{center}
    \includegraphics[width=1\textwidth]{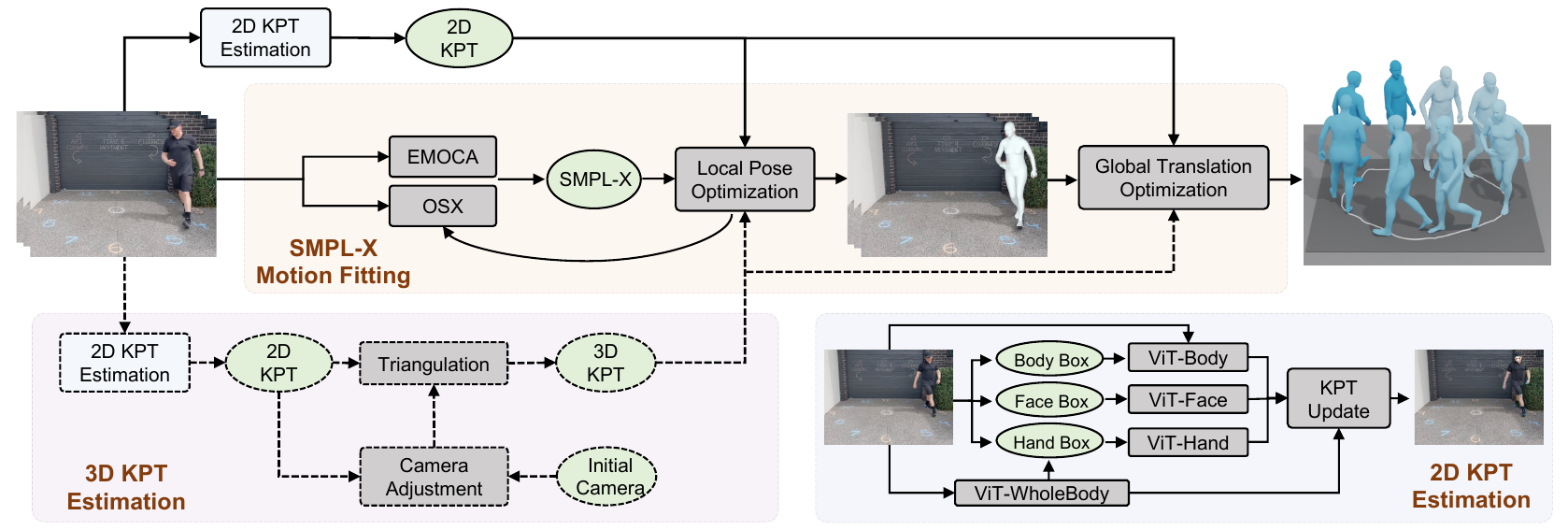}
\end{center}
\vspace{-0.1cm}
\caption{\small The automatic pipeline for the whole-body motion capture from  massive videos, including 2D and 3D whole-body keypoints estimation, local pose optimization, and global translation optimization. This pipeline supports both single- and multi-view inputs. Dashed lines represent the handling of multi-view data exclusively.}

\label{figure:motion_anno}
\vspace{-0.4cm}
\end{figure*}

\noindent\textbf{Overview.}
To efficiently capture a large volume of potential motions from massive videos, we propose an annotation pipeline for high-quality whole-body motion capture with 
\emph{three novel techniques}: 
(i) hierarchical whole-body keypoint estimation; 
(ii) score-guided adaptive temporal smoothing for jitter motion refinement; 
and (iii) learning-based 3D human model fitting for accurate motion capture. 

\noindent\textbf{2D Keypoint Estimation.} 2D Whole-body keypoint estimation poses a challenge due to the small size of the hands and face regions.
Although recent approaches have utilized separate networks to decode features of different body parts~\cite{coco_wholebody,xu2022zoomnas}, 
they often struggle with hand-missing detection and are prone to errors due to occlusion or interaction.
To overcome these limitations, we customize a novel hierarchical keypoint annotation method, depicted in the blue box of Fig.~\ref{figure:motion_anno}. 
We train a ViT-WholeBody based on a ViT-based model~\cite{YufeiXu2022ViTPoseSV} on the COCO-Wholebody dataset~\cite{coco_wholebody} to estimate initial whole-body 
keypoints $\mathbf{K}^{\text{2D}}\in\mathbb{R}^{133\times2}$ with confidence scores. 
Leveraging the ViT model's ability to model semantic relations between full-body parts, 
we enhance hand and face detection robustness even under severe occlusion.
Subsequently, we obtain the hand and face bounding boxes based on the keypoints, and refine the boxes using the BodyHands detector~\cite{bodyhands} through an IoU matching operation. 
Finally, we feed the cropped body, hand, and face regions into three separately pre-trained ViT networks to estimate body, hand and face keypoints, which are used to update $\mathbf{K}^{\text{2D}}$.

\noindent\textbf{Score-guided Adaptive Smoothing.} To address the jitter resulting from per-frame pose estimation 
in challenging scenarios such as heavy occlusion, truncation, and motion blur, while preserving motion details, 
we introduce a novel score-guided adaptive smoothing technique into the traditional Savitzky-Golay filter~\cite{savitzky1964smoothing}. 
The filter is applied to a sequence of 2D keypoints of a motion:
\begin{equation}
\small
    \bar{\mathbf{K}}_i^{\text{2D}} = \sum_{j=-w}^w c_j \mathbf{K}^{\text{2D}}_{i+j},
\end{equation}
where ${\mathbf{K}}_i^{\text{2D}}$ is the original keypoints of the $i_\text{th}$ frame, 
$\bar{\mathbf{K}}_i^{\text{2D}}$ is the smoothed keypoints,
$w$ corresponds to half-width of filter window size, 
and $c_j$ are the filter coefficients. 
Different from existing smoothing methods with a fixed window size~\cite{zeng2022deciwatch,zeng2022smoothnet,savitzky1964smoothing}, we leverage the confidence scores of the keypoints to adaptively adjust the window size to balance between smoothness and motion details. 
Using a larger window size for keypoints with lower confidence scores can mitigate the impact of outliers.

\noindent\textbf{3D Keypoint Annotation.} Precise 3D keypoint can boost the estimation of SMPL-X. We utilize novel information from large-scale pre-trained models. Accordingly, for single-view videos, we adopt a pretrained model~\cite{sarandi2023learning}, which is trained on massive 3D datasets, to estimate precise 3D keypoints. 
For multi-view videos, we utilize bundle adjustment to calibrate and refine the camera parameters, and then triangulate the 3D keypoints $\bar{\mathbf{K}}^{\text{3D}}$ based on the multi-view 2D keypoints. To enhance stability, we adopt temporal smoothing and enforce 3D bone length constraints during triangulation.

\noindent\textbf{Local Pose Optimization.} After obtaining the keypoints, we perform local pose optimization to register each frame's whole-body model SMPL-X~\cite{smpl-x}. Traditional optimization-based methods~\cite{smplify, smpl-x} are often time-consuming and may yield unsatisfactory results as they ignore image clues and motion prior. We propose a progressive learning-based human mesh fitting method to address these limitations. Initially, we predict the SMPL-X parameter $\Theta$
with the SOTA whole-body mesh recovery method OSX~\cite{osx} and face reconstruction model EMOCA~\cite{emoca}. 
And then, through iterative optimization of the network parameters, we fit the human model parameters $\hat{\Theta}$ to the target 2D and 3D joint positions by minimizing the following functions, achieving an improved alignment accuracy:
\begin{equation}
\small
    L_\text{joint} = \Vert \hat{\mathbf{K}}^\text{3D}-\bar{\mathbf{K}}^\text{3D} \Vert_1 + \Vert \hat{\mathbf{K}}^\text{2D}-\bar{\mathbf{K}}^\text{2D} \Vert_1
     + \Vert \hat{\Theta} - \Theta \Vert_1.
\end{equation}
Here, $\hat{\mathbf{K}}^\text{3D}$ represents the predicted 3D joint positions obtained by applying a linear regressor to a 3D mesh generated by the SMPL-X model. $\hat{\mathbf{K}}^\text{2D}$ is derived by performing a perspective projection of the 3D keypoints. The last term of the loss function provides explicit supervision based on the initial parameter, serving as a 3D motion prior. 
To alleviate potential biophysical artifacts, such as interpenetration and foot skating, we incorporate a set of physical optimization constraints:
\begin{equation}
\small
    L = \lambda_\text{joint}L_\text{joint} + \lambda_\text{smooth}L_\text{smooth} + \lambda_\text{pen}L_\text{pen} + \lambda_\text{phy}L_\text{phy}.
\end{equation}
Here, $\lambda$ are weighting factors of each loss function and $L_\text{smooth}$ is a first-order smoothness term:
\begin{equation}
\small
    L_\text{smooth} = \sum_t \Vert \hat{\Theta}_{2:t}-\hat{\Theta}_{1:t-1} \Vert_1 + \sum_t \Vert \hat{\mathbf{K}}^\text{3D}_{2:t}-\hat{\mathbf{K}}^\text{3D}_{1:t-1} \Vert_1,
\end{equation}
where $\hat{\Theta}_i$ and $\hat{\mathbf{K}}^\text{3D}_i$ represent the SMPL-X parameters and joints of the $i$-th frame, respectively. To alleviate mesh interpenetration, we utilize a collision penalizer~\cite{collision}, denoted as $L_{\text{pen}}$. Additionally, we incorporate the physical loss $L_\text{phy}$ based on PhysCap~\cite{physcap} to prevent implausible poses.

\noindent\textbf{Global Motion Optimization.} To improve the  consistency and realism of the estimated global trajectory,
we perform a global motion optimization based on GLAMR~\cite{glamr} to simultaneously refine the global motions and camera poses to align with video evidence, such as 2D keypoints:
\begin{equation}
\small
    L_g = \lambda_\text{2D}L_\text{2D} + \lambda_\text{traj}L_\text{traj} + \lambda_\text{cam}L_\text{cam} + \lambda_\text{reg}L_\text{reg},
\end{equation}
where $L_\text{2D}$ represents the 2D keypoint distance loss, $L_\text{traj}$ quantifies the difference between the optimized global trajectory and the trajectory estimated by Kama~\cite{kama}. $L_\text{reg}$ enforces regularization on the global trajectory, and $L_\text{cam}$ applies a smoothness constraint on the camera parameters.

\noindent\textbf{Human Verification.} To ensure quality, we manually checked the annotation by removing the motions that do not align with the video evidence or exhibit obvious biophysical artifacts.

\subsection{Obtaining Whole-body Motion Descriptions}
\label{sec:lable_text}
\vspace{-0.2cm}

\noindent\textbf{Sequence motion labels.} 
The videos in \dataname were collected from online sources and existing datasets. 
For action-related datasets~\cite{cai2022humman,chung2021haa500,li2021aist,liu2019ntu,amass,taheri2020grab}, we use the action labels as one of the sequence semantic labels. Meanwhile, we input the videos into Video-LLaMA~\cite{damonlpsg2023videollama} and filter the human action descriptions as supplemental texts. 
When videos contain semantic subtitles, EasyOCR automatically extracts semantic information. For online videos, we also use the search queries generated from LLM~\cite{jiao2023chatgpt} as semantic labels.
Videos without available semantic information, such as EgoBody~\cite{zhang2022egobody}, are manually labeled using the VGG Image Annotator (VIA)~\cite{dutta2019vgg}.
For the face database BAUM~\cite{baum}, we use the facial expression labels provided by the original creator.

\begin{figure*}[h]
\vspace{-0.3cm}
\begin{center}
    \includegraphics[width=1\textwidth]{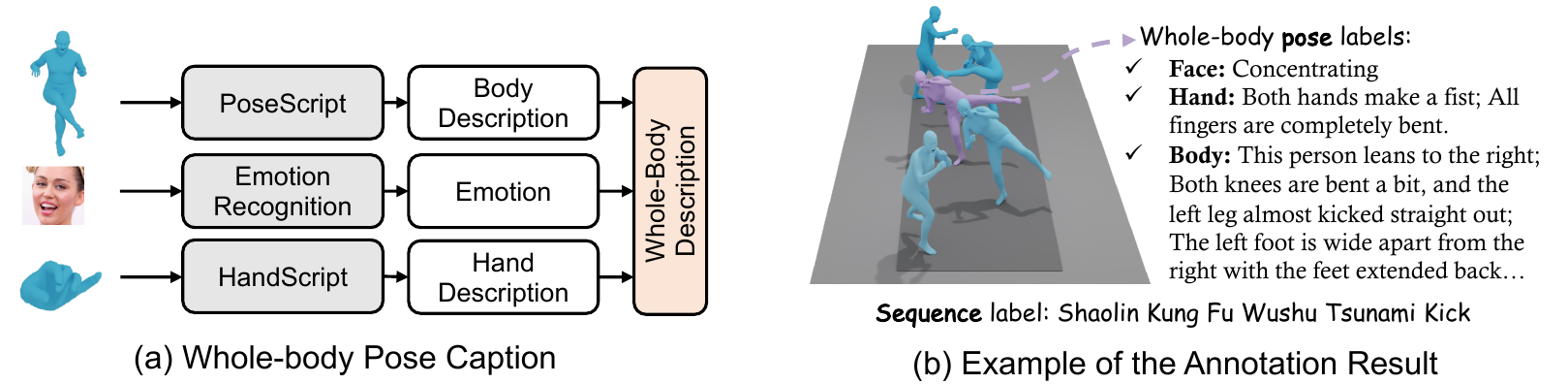}
\end{center}
\vspace{-0.3cm}
\caption{\small Illustration of (a) annotation of the whole-body pose description, and (b) an example of the text labels.}
\label{figure:text_data_collection}
\vspace{-0.3cm}
\end{figure*}

\noindent\textbf{Whole-body pose descriptions.} The generation of fine-grained pose descriptions for each pose involves three distinct parts: face, body, and hand, as shown in Fig.~\ref{figure:text_data_collection}(a).
\emph{Facial expression labeling} uses the emotion recognition model EMOCA~\cite{emoca} pretrained on AffectNet~\cite{mollahosseini2017affectnet} to classify the emotion. 
\emph{Body-specific descriptions} utilizes the captioning process from PoseScript~\cite{posescript}, which generates synthetic low-level descriptions in natural language based on given 3D keypoints. The unit of this information is called posecodes, such as \textit{`the knees are completely bent'}. A set of generic rules based on fine-grained categorical relations of the different body parts are used to select and aggregate the low-level pose information.  The aggregated posecodes are then used to produce textual descriptions in natural language using linguistic aggregation principles.
\emph{Hand gesture descriptions} extends the pre-defined posecodes from body parts to fine-grained hand gestures. 
We define six elementary finger poses via finger curvature degrees and distances between fingers to generate descriptions, such as \textit{`bent'} and \textit{`spread apart'}. 
We calculate the angle of each finger joint based on the 3D hand keypoints and determine the corresponding margins.
For instance, if the angle between $\vec{\mathbf{V}}(\mathbf{K}_\text{wrist}, \mathbf{K}_\text{fingertip})$ and $\vec{\mathbf{V}}(\mathbf{K}_\text{fingertip}, \mathbf{K}_\text{fingeroot})$ falls between 120 and 160 degrees, the finger posture is labeled as \textit{`slightly bent'}.
We show an example of the annotated text labels in Fig.~\ref{figure:text_data_collection}(b). 

\textbf{Summary.} Based on the above annotations, we bulid \dataname, which has 81.1K clips with 15.6M SMPL-X poses and the corresponding pose and semantic text labels.

\vspace{-0.3cm}
\section{Experiment}
\vspace{-0.3cm}
In this section, we first validate the accuracy of our motion annotation pipeline on the 2D keypoints and 3D SMPL-X datasets.
Then, we build a text-driven whole-body motion generation benchmark on \dataname. 
Finally, we show the effectiveness of \dataname in whole-body human mesh recovery. 
\vspace{-0.2cm}

\begin{figure*}[h]
\vspace{-0.1cm}
\begin{center}
    \includegraphics[width=0.97\textwidth]{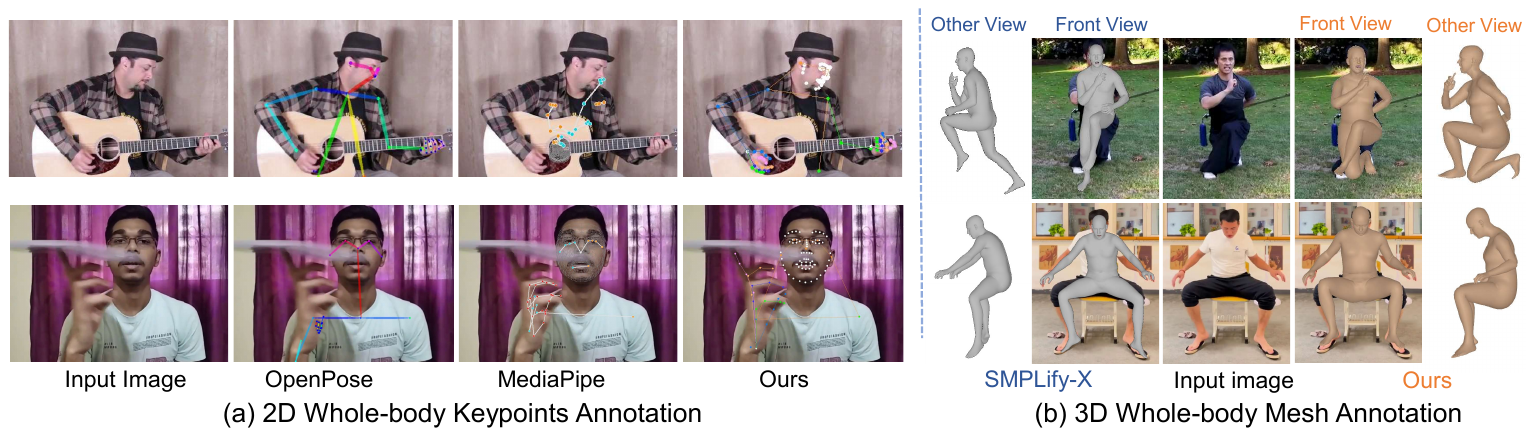}
\end{center}
\vspace{-0.4cm}
\caption{\small Qualitative comparisons of (a) 2D keypoints annotation with widely used methods~\cite{openpose, mediapipe} and (b) the 3D mesh annotation with the popular fitting method~\cite{smpl-x} with ours.}
\label{figure:motion_anno_compare}
\vspace{-0.8cm}
\end{figure*}

\begin{figure}[h]
  \centering
    \subfloat[\small Evaluation result on COCO-Wholebody~\cite{coco_wholebody} dataset.]{
      \scalebox{0.66}{
      \begin{tabular}{l|cc|cc|cc}
        \toprule
        \rowcolor{lightgray}
        & \multicolumn{2}{c|}{hand $\uparrow$} 
        & \multicolumn{2}{c|}{face $\uparrow$} 
        & \multicolumn{2}{c}{whole-body $\uparrow$} \\
        \rowcolor{lightgray}
        \multirow{-2}{*}{{Method}} &
        {AP} & {AR} & {AP} & {AR} & {AP} & {AR} \\
        \midrule
        OpenPose~\cite{openpose} & 38.6 & 43.3 & 76.5 & 84.0 & 44.2 & 52.3  \\
        HRNet~\cite{hrnet}  & 50.3 & 60.2 & 73.7 & 80.9 & 58.2 & 67.1  \\
        ViTPose~\cite{YufeiXu2022ViTPoseSV}  & 47.4 & 59.4 & 59.8 & 70.7 & 57.7 & 69.4  \\
        RTMPose-l~\cite{rtmpose}  & 52.3 & 60.0 & 84.4 & 87.6 & 63.2 & 69.4  \\
        Ours & \textbf{64.9} & \textbf{74.0} & \textbf{91.6} & \textbf{94.4} 
        & \textbf{73.5} {\color{red}$\uparrow_{16.3\%}$} & \textbf{80.3} {\color{red}$\uparrow_{15.7\%}$} \\
        \bottomrule
        \end{tabular}}
      }\hfill
    \subfloat[\small Reconstruction error on EHF~\cite{smpl-x} dataset.]{
      \scalebox{0.64}{
        \begin{tabular}{lccc}
          \toprule
          \rowcolor{lightgray}
          Method & PA-MPJPE $\downarrow$ & PA-MPVPE $\downarrow$ & MPVPE $\downarrow$  \\
          \midrule
          Hand4Whole~\cite{GyeongsikMoon2020hand4whole} & 58.9 & 50.3 & 79.2 \\
          OSX~\cite{osx} & 55.6& 48.7 & 70.8 \\
          PyMAF-X~\cite{pymafx} & 52.8 & 50.2 & 64.9 \\
          SMPLify-X~\cite{smpl-x} & 62.6 & 52.9 & - \\
          \midrule
          Ours & 33.5 & 31.8 & 44.7{\color{red}$\downarrow_{30.1\%}$} \\
          Ours w/GT 3Dkpt & \textbf{23.9} & \textbf{19.7} &  \textbf{30.7} {\color{red}$\downarrow_{52.7\%}$}\\
          \bottomrule
          \end{tabular}}   
    }
    \vspace{-0.2cm}
  \captionof{table}{\small Evaluation of motion annotation pipeline on (a) 2D keypoints and (b) 3D SMPL-X datasets.}
  \label{tab:annot_eval}
  \vspace{-0.4cm}
  \end{figure}

\subsection{Evaluation of the Motion Annotation Pipeline}
\label{sec:exp_eval}
\vspace{-0.2cm}
\textbf{2D Keypoints Annotation.} We evaluate the proposed 2D keypoint annotation method on the COCO-WholeBody~\cite{coco_wholebody} dataset, 
and compare the evaluation result with four SOTA keypoints estimation methods ~\cite{openpose, hrnet, YufeiXu2022ViTPoseSV, rtmpose}.
We use the same input image size of $256\times 192$ for all the methods to ensure a fair comparison. 
From Tab.~\ref{tab:annot_eval}(a), our annotation pipeline significantly surpasses existing methods by over 15\% average precision. Additionally, we provide qualitative comparisons in Fig.~\ref{figure:motion_anno_compare}(a), illustrating the robust and superior performance of our method, especially in challenging and occluded scenarios. 

\textbf{3D SMPL-X Annotation.}
We evaluate our learning-based fitting method on the EHF~\cite{smpl-x} dataset and compare it with four open-sourced human mesh recovery methods. Following previous works,
we employ mean per-vertex error (MPVPE), Procrusters aligned mean per-vertex error (PA-MPVPE), and Procrusters aligned mean per-joint error (PA-MPJPE) as evaluation metrics (in mm).
Results in Tab.~\ref{tab:annot_eval}(b) demonstrate the superiority of our progressive fitting method (over 30\% error reduction). 
Specifically, PA-MPVPE is only 19.71 mm when using ground-truth 3D keypoints as supervision. 
Fig.~\ref{figure:motion_anno_compare}(b) shows the annotated mesh from front and side view, indicating reliable 3D SMPL-X annotations with reduced depth ambiguity. More results are presented in Appendix due to page limits.

\vspace{-0.2cm}
\subsection{Impact on Text-driven Whole-body Motion Generation}
\label{sec:t2m_exp}
\vspace{-0.2cm}
\noindent\textbf{Experiment Setup.} We randomly split \dataname   into the train ($80\%$), val ($5\%$), and test ($15\%$) sets. SMPL-X is adopted as the motion representation for expressive motion generation.

\noindent\textbf{Evaluation metrics.}
We adopt the same evaluation metrics as \cite{humanml3d}, including Frechet Inception Distance (FID), Multimodality, Diversity, R-Precision, and Multimodal Distance. 
Due to the page limit, we leave more details about experimental setups and evaluation metrics in the appendix.

\begin{table*}[t]
	\vspace{4pt}
        \centering
	\resizebox{1.0\textwidth}{!}{%
		\begin{tabular}{lccccccc}
			\toprule
      \rowcolor{color3}
			 & \multicolumn{3}{c}{R Precision $\uparrow$}& & & & 
       \\ 
      \rowcolor{color3}
			\multirow{-2}{*}{Methods} & \multicolumn{1}{c}{Top 1} & 
      \multicolumn{1}{c}{Top 2} & 
      \multicolumn{1}{c}{Top 3} & 
      \multirow{-2}{*}{FID$\downarrow$} & 
      \multirow{-2}{*}{MM Dist$\downarrow$} & 
      \multirow{-2}{*}{Diversity$\rightarrow$} & 
      \multirow{-2}{*}{MModality}
      \\ 
      \midrule
			Real &
			$0.573^{\pm.005}$ &
			$0.765^{\pm.003}$ &
			$0.850^{\pm.005}$ &
			$0.001^{\pm.001}$&
			$2.476^{\pm.002}$&
			$13.174^{\pm.227}$ &
			- \\
			\midrule
			MDM \cite{Tevet2022HumanMD}&
                $0.290^{\pm.011}$ &
                $0.459^{\pm.010}$ &
                $0.577^{\pm.008}$ &
                $2.094^{\pm.230}$ &
                $6.221^{\pm.115}$ &
                $11.895^{\pm.354}$ &
                $2.624^{\pm.083}$
                \\
      MLD \cite{mld}&
			$0.440^{\pm.002}$ &
			$0.624^{\pm.004}$ &
			$0.733^{\pm.003}$ &
			$0.914^{\pm.056}$ &
			$3.407^{\pm.020}$ &
			$13.001^{\pm.245}$ &
			$2.558^{\pm.084}$
                \\ 
                
      T2M-GPT \cite{t2m-gpt}&
      $0.502^{\pm.004}$ &
      $0.697^{\pm.005}$ &
      $0.791^{\pm.007}$ &
      $0.699^{\pm.012}$ &
      $3.192^{\pm.035}$ &
      $\boldsymbol{13.132}^{\pm.127}$ &
      $2.510^{\pm.027}$ \\ 
      
        MotionDiffuse \cite{motiondiffuse} &
        $\boldsymbol{0.559}^{\pm.001}$ &
        $\boldsymbol{0.748}^{\pm.004}$ &
        $\boldsymbol{0.842}^{\pm.003}$ &
        $\boldsymbol{0.457}^{\pm.007}$ &
        $\boldsymbol{2.542}^{\pm.018}$ &
        $13.576^{\pm.161}$ &
        ${1.620}^{\pm.152}$ \\

			  \bottomrule
		\end{tabular}%
	}
	\vspace{-0.2cm}
	\caption{\small Benchmark of text-driven motion generation on \dataname test set. 
  `$\rightarrow$' means results are better if the metric is closer to the real motions and $\pm$ indicates the $95\%$ confidence interval. The best results are in \textbf{bold}.} 
	\label{tab:benchmark}
\vspace{-0.3cm}
\end{table*}


\begin{table*}[t]
\centering
\resizebox{1.0\textwidth}{!}{
\begin{tabular}{l|cccc|cccc}
  \toprule
  \rowcolor{lightgray}
  & 
  \multicolumn{4}{c|}{HumanML3D (Test)} & 
  \multicolumn{4}{c}{Motion-X (Test)} \\
  \rowcolor{lightgray}
  \multicolumn{1}{c|}{\multirow{-2}{*}{{Train Set}}} &
  {R-Precision$\uparrow$} & 
  {FID$\downarrow$} & 
  {Diversity$\rightarrow$} & 
  {MModality}& 
  {R-Precision$\uparrow$} & 
  {FID$\downarrow$} & 
  {Diversity$\rightarrow$} & 
  {MModality} \\
  \midrule
  Real (GT) & $0.749^{\pm.002}$ & $0.002^{\pm.001}$ & $9.837^{\pm.084}$ & - & $0.850^{\pm.005}$ & $0.001^{\pm.001}$ &  $13.174^{\pm.227}$ & -  \\
  \midrule
  HumanML3D & $0.657^{\pm.004}$ & $1.579^{\pm.050}$ & $10.098^{\pm.052}$ & $2.701^{\pm.143}$ & $0.570^{\pm..003}$ & $12.309^{\pm.127}$ & $9.529^{\pm.165}$ & $2.960^{\pm.066}$ \\
  Motion-X &  $\mathbf{0.695}^{\pm.005}$ & $\mathbf{0.999}^{\pm.042}$ & $\mathbf{9.871}^{\pm.099}$ & $\mathbf{2.827}^{\pm.138}$ & $\mathbf{0.733}^{\pm.003}$ & $\mathbf{0.914}^{\pm.056}$ & $\mathbf{13.001}^{\pm.245}$ & $2.558^{\pm.084}$ \\ 
  \bottomrule
  \end{tabular}}
  \vspace{-0.2cm}
  \caption{\small Cross-dataset comparisons of HumanML3D and \dataname. We train MLD on the training set of HumanML3D and \dataname, respectively, then evaluate it on their test sets.}
  \label{tab:compare_humanml}
  \vspace{-0.6cm}
\end{table*}

\noindent\textbf{Benchmarking Motion-X.} We train and evaluate four diffusion-based motion generation methods, including MDM~\cite{Tevet2022HumanMD}, MLD~\cite{mld}, MotionDiffuse~\cite{motiondiffuse} and T2M-GPT~\cite{t2m-gpt} on our dataset. Since previous datasets only have sequence-level motion descriptions, we keep similar settings for minimal model adaptation and take the semantic label as text input. The evaluation is conducted with 20 runs (except for Multimodality with 5 runs) under a $95\%$ confidence interval. From Tab.~\ref{tab:benchmark}, MotionDiffuse demonstrates a superior performance across most metrics. However, it scores the lowest in Multimodality, indicating that it generates less varied motion. Notably, T2M-GPT achieves comparable performance on our dataset while maintaining high diversity, indicating our large-scale dataset's promising prospects to enhance the GPT-based method's efficacy. MDM gets the highest Multimodality score with the lowest precision, indicating the generation of noisy and jittery motions. The highest Top-1 precision is 55.9\%, showing the challenges of \dataname.
MLD adopts the latent space design, making it fast while maintaining competent results. Therefore, we use MLD to conduct the following experiments to compare \dataname with the existing largest motion dataset HumanML3D and ablation studies.

\begin{wrapfigure}{r}{0.50\textwidth}
  \centering
  \vspace{-0.5cm}
  \includegraphics[width=0.50\textwidth]{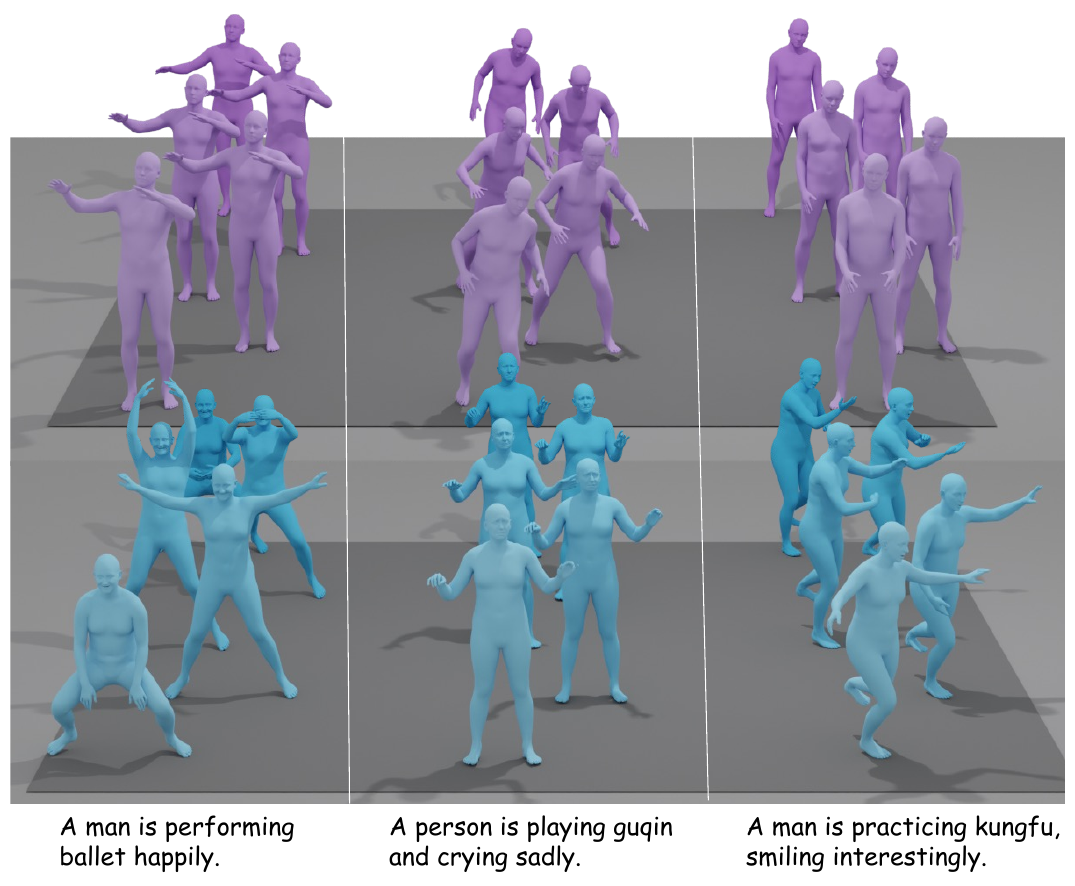}
  \vspace{-0.6cm}
  \caption{\small Visual comparisons of motions generated by MLD~\cite{mld} trained on HumanML3D (in purple) or \dataname (in blue). Please zoom in for a detailed comparison. The model trained with \dataname can generate more accurate and semantic-corresponded motions.}
  \label{fig:motion_gen}
  \vspace{-0.3cm}
\end{wrapfigure}

\noindent \textbf{Comparison with HumanML3D.} To validate the richness, expressiveness, and effectiveness of our dataset, we conduct a comparative analysis between \dataname and HumanML3D, which is the largest existing dataset with text-motion labels. We replace the original vector-format poses of HumanML3D with the corresponding SMPL-X parameters from AMASS~\cite{amass}, and randomly extract facial expressions from BAUM~\cite{baum} to fill in the face parameters. We train MLD separately on the training sets of \dataname and HumanML3D, then evaluate both models on the two test sets. The results in Tab.~\ref{tab:compare_humanml} reveal some valuable insights. Firstly, \dataname exhibits greater diversity (\textbf{13.174}) than HumanML3D (\textbf{9.837}), as evidenced by the real (GT) row. This indicates a wider range of motion types captured by \dataname. Secondly, the model pretrained on \dataname and then finetuned on HumanML3D subset performs well on the HumanML3D test set, even better than the intra-data training. These superior performances stem from the fact that \dataname encompasses diverse motion types from massive outdoor and indoor scenes.  For a more intuitive comparison, we provide the visual results of the generated motion in Fig.~\ref{fig:motion_gen}, where we can clearly see that the model trained on \dataname excels at synthesizing semantically corresponding motions given text inputs.
These results prove the significant advantages of \dataname in enhancing expressive, diverse, and natural motion generation. 

\noindent\textbf{Ablation study of text labels.}
In addition to sequence-level semantic labels, the text labels in \dataname also include frame-level pose descriptions, which is an important characteristic of our dataset. To assess the effectiveness of pose description, we conducted an ablation study on the text labels. The baseline model solely utilizes the semantic label as the text input. Since there is no method to use these labels, we simply sample a single sentence from the pose descriptions randomly, concatenate it with the semantic label, and feed the combined input into the CLIP text encoder. Interestingly, from Tab.~\ref{tab:ablation}, adding additional face and body pose texts brings consistent improvements, 
and combining whole-body pose descriptions results in a noteworthy $38\%$ reduction in FID.
These results validate that the proposed whole-body pose description contributes to generating more accurate and realistic human motions. More effective methods to utilize these labels can be explored in the future.

\begin{figure}[t]
  \begin{minipage}{0.44\textwidth}
  \centering
  \vspace{-0.3cm}
  \resizebox{\linewidth}{!}{
      \makeatletter\def\@captype{table}\makeatother
      \begin{tabular}{c c c c c}
        \toprule
        \rowcolor{lightgray}
        Semantic & \multicolumn{3}{c}{Pose Discription} 
        &  \\
        \rowcolor{lightgray}
        Label & face text & body text & hand text  & 
        \multirow{-2}{*}{~FID$\downarrow$~} \\
        \midrule
        \checkmark &  &  &   & $0.914^{\pm.056}$ \\
        \checkmark &\checkmark  &  &  & $0.784^{\pm.032}$ \\
        \checkmark  & \checkmark & \checkmark &  & $0.671^{\pm.016}$ \\
        \checkmark  &\checkmark  & \checkmark & \checkmark & $\mathbf{0.565}^{\pm.036}$ \\
        \bottomrule
        \end{tabular}
      }
      \vspace{-0.1cm}
      \captionof{table}{\small Ablation study of text inputs.}
      \label{tab:ablation}
  \end{minipage}
  \hfill
  \begin{minipage}{0.55\textwidth}
    \centering
    \vspace{-0.5cm}
    \resizebox{\linewidth}{!}{
        \makeatletter\def\@captype{table}\makeatother
        \begin{tabular}{l|ccc|ccc}
          \toprule
          \rowcolor{lightgray}
          & \multicolumn{3}{c|}{\boldmath{}{EHF~\cite{smpl-x} $\downarrow$}\unboldmath{}} & \multicolumn{3}{c}{\boldmath{}{AGORA~\cite{action2video} $\downarrow$}\unboldmath{}} \\
          \rowcolor{lightgray}
          \multicolumn{1}{c|}{\multirow{-2}{*}{{Method}}} & {all} & {hand} & {face} & {all} & {hand} & {face} \\
          \midrule
          w/o Motion-X & 79.2 & 43.2 &25.0 &185.6 & 73.7 & 82.0   \\
          w/ Motion-X & \textbf{73.0} & \textbf{41.0} & \textbf{22.6} &\textbf{184.1} & \textbf{73.3} & \textbf{81.4}  \\
          \bottomrule
          \end{tabular}
        }
        \vspace{-0.1cm}
        \captionof{table}{\small Mesh recovery errors of Hand4Whole~\cite{GyeongsikMoon2020hand4whole}
        using different training datasets. MPVPE (mm) is reported.}
        \vspace{-0.3cm}
        \label{tab:mesh_recovery}
    \end{minipage}
\end{figure}

\subsection{Impact on Whole-body Human Mesh Recovery}
\label{sec:mesh_recovery}
\vspace{-0.2cm}
As discovered in this benchmark~\cite{pang2022benchmarking}, the performance of mesh recovery methods can be significantly improved by utilizing high-quality pseudo-SMPL labels. \dataname provides a large volume of RGB images and well-annotated SMPL-X labels.
To verify its usefulness in the 3D whole-body mesh recovery task, we take Hand4Whole~\cite{GyeongsikMoon2020hand4whole} as an example and evaluate MPVPE on the widely-used AGORA val~\cite{agora} and EHF~\cite{smpl-x} datasets. For the baseline model, we train it on the commonly used COCO~\cite{coco_wholebody}, Human3.6M~\cite{hm36}, and MPII~\cite{andriluka2014mpii} datasets. We then train another model by incorporating an additional $10\%$ of the single-view data sampled from \dataname while keeping the other setting the same. As shown in Tab.~\ref{tab:mesh_recovery}, the model trained with \dataname shows a significant decrease of $7.8\%$ in MPVPE on EHF and AGORA compared to the baseline model. The gains come from the increase in diverse appearances and poses in \dataname, indicating the effectiveness and accuracy of the motion annotations in \dataname and its ability to benefit the 3D reconstruction task.

\section{Conclusion}
In this paper, we present \emph{Motion-X}, a comprehensive and large-scale 3D expressive whole-body human motion dataset. It addresses the limitations of existing mocap datasets, which primarily focus on indoor body-only motions with limited action types. The dataset consists of 144.2 hours of whole-body motions and corresponding text labels.
To build the dataset, we develop a systematic annotation pipeline to  annotate 81.1K 3D whole-body motions, sequence-level motion semantic labels, and 15.6M frame-level whole-body pose descriptions.
Comprehensive experiments demonstrate the accuracy of the motion annotation pipeline and the significant benefit of \dataname in enhancing expressive, diverse, and natural motion generation, as well as 3D whole-body human mesh recovery. 

\textbf{Limitation and future work.} 
There are two main limitations of our work. Firstly, the motion quality of our markless motion annotation pipeline is inevitably inferior to the multi-view marker-based motion capture system. Secondly, during our experiment, we found out that existing evaluation metrics are not always consistent with visual results. Thus, there is a need for further development of the motion generation models and evaluation metrics.
As a large-scale dataset with multiple modalities, e.g., motion, text, video, and audio, \emph{Motion-X} holds great potential for advancing downstream tasks, such as motion prior learning, understanding, and multi-modality pre-training. Besides, our large-scale dataset and scalable annotation pipeline open up possibilities for combining this task with a large language model (LLM) to achieve an exciting motion generation result in the future.
With \emph{Motion-X}, we hope to benefit and facilitate further research in relevant fields. 

\section*{Acknowledgements}

This research is partially funded through National Key Research and Development Program of China (Project No. 2022YFB36066), in part by the Shenzhen Science and Technology Project under Grant (JCYJ20220818101001004, JSGG20210802153150005), National Natural Science Foundation of China (62271414), the Young Scientists Fund of the National Natural Science Foundation of China under grant No.62106154, the Natural Science Foundation of Guangdong Province, China (General Program) under grant No.2022A1515011524, and Shenzhen Science and Technology Program JCYJ20220818103001002 and Shenzhen Science and Technology Program ZDSYS20211021111415025. 


Thanks to everyone who participated in shooting the IDEA400 dataset and discussions. Thanks to Linghao Chen for helping to design the icons and website pages.
\clearpage

\appendix

\section*{Appendix}

\section{Motion-X: Additional Details}

In this section, we provide more details about \dataname that are not included in the main paper due to space limitations, including statistic analyses, data preprocessing, and motion augmentation mechanism.

\subsection{Statistic Analyses}

\begin{figure}[h]	
    \centering
    \vspace{-0.2cm}
        {
            \begin{minipage}[t]{0.4\linewidth}
                \centering      
                \includegraphics[width=2.4in]{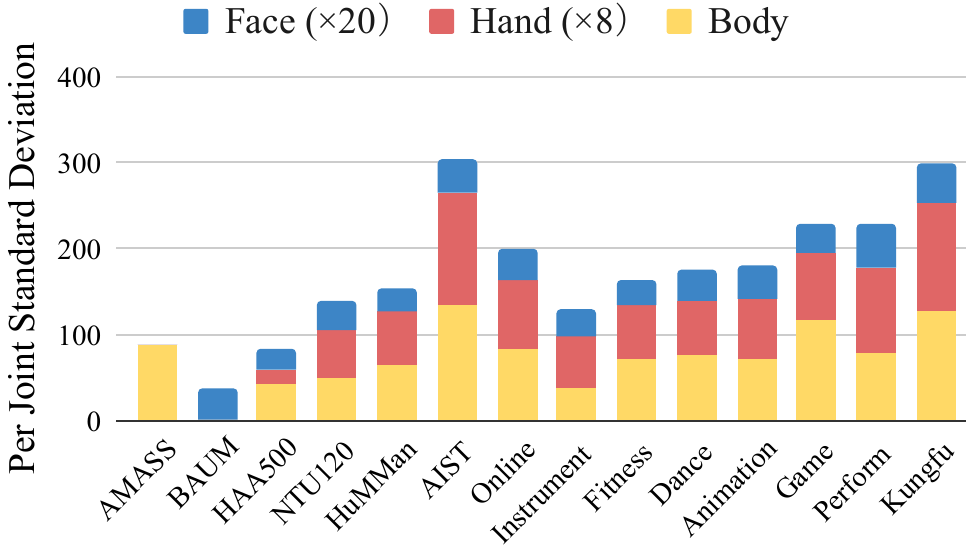}
            \end{minipage}
        }
            \label{fig:std_data}  
            \hspace{1cm}
        {
            \begin{minipage}[t]{0.4\linewidth}
                \centering         
                \includegraphics[width=1.9in]{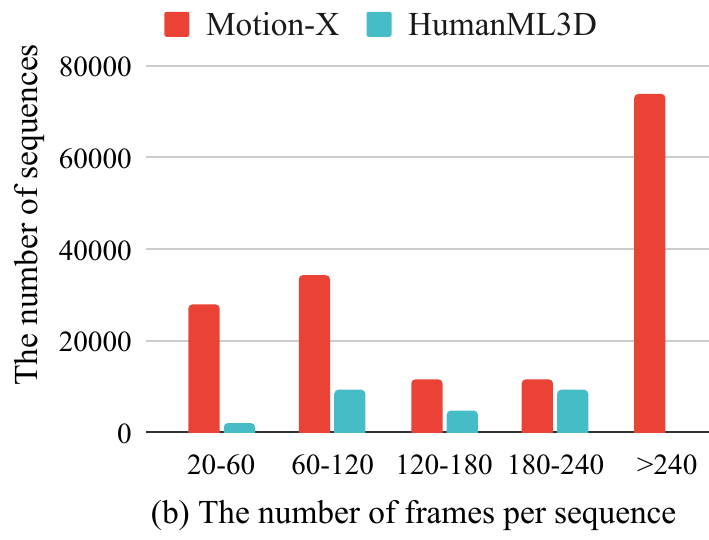}   
            \end{minipage}
        } 
            \label{fig:frame_data} 
    \vspace{0.2cm}
    \caption{Statistics of motion diversity and length.}
    \vspace{-0.4cm}
    \label{fig:data_comp_sup} 
    \end{figure}

Fig.~\ref{fig:data_comp_sup}(a) shows each sub-dataset standard deviation of body, hand, and face joints. Our dataset has a large diversity of the hand and face joints, filling the gap of the previous body-only dataset in terms of expressiveness. Besides, as shown in Fig.~\ref{fig:data_comp_sup}(b), \dataname provides a large volume of long motion ($>240$ frames), which will be beneficial for long-term motion generation.

\subsection{Processing of Each Sub-dataset}

We gather 81.1K motion sequences from seven existing datasets and a large volume of online videos with the proposed annotation pipeline. As shown in Tab.~\ref{tab:statistic_subdataset}, due to the lack of comprehensive annotations from their original datasets, we provide well-annotated whole-body motion, comprehensive semantic labels, and whole-body pose descriptions for all datasets. Here we introduce more details about each sub-dataset's data preprocessing.

\begin{table*}[h]
	\vspace{4pt}
        \centering
	\resizebox{0.9\textwidth}{!}{%
    \begin{tabular}{lccccc}
        \toprule
            \rowcolor{color3} \text{Dataset} &~~~\text{Clip Number}~~~ & ~~\text{Frame Number}~~ & \text{Motion Annotation} & \text{Text Annotation} & \text{RGB} \\        \midrule
            AMASS~\cite{humanml3d} & 26.3K & 5.4M & B,H & S & \xmark \\ 
            AIST~\cite{tsuchida2019aist} & 1.4K & 0.3M & B & S &  \cmark\\
            HAA500~\cite{chung2021haa500} & 5.2K & 0.3M & - & S &  \cmark \\ 
            HuMMan~\cite{cai2022humman} & 0.7K & 0.1M & B & S &  \cmark \\
            GRAB~\cite{taheri2020grab} & 1.3K & 0.4M & B,H & S  & \xmark\\ 
            EgoBody~\cite{zhang2022egobody} & 1.0K & 0.4M & B,H & - &  \cmark \\ 
            BAUM~\cite{baum} & 1.4K & 0.2M & F & S  & \cmark\\ 
            IDEA400* & 12.5K & 2.6M & - & - &  \cmark \\
            Online Videos* & 32.5K & 6.0M & - & - &  \cmark\\ \midrule
            Motion-X & 81.1K & 15.6M & B,H,F & S,P & \cmark \\ \bottomrule
        \end{tabular}
	}
	\caption{\small Statistics of sub-datasets. B, H, and F are body, hand, and face. S and P are semantic and pose texts. Please note that the semantic text annotations are different from different datasets. Some action datasets~\cite{liu2019ntu,amass,chung2021haa500,baum} only provide action-level texts instead of sentences. Most semantic labels are annotated manually, making specific domain descriptions (e.g., specific instruments) hard to label precisely.  * denotes videos are collected in this work.}
	\label{tab:statistic_subdataset}
\vspace{-0.3cm}
\end{table*}

\textbf{AMASS}~\cite{amass} is the existing largest-scale motion capture dataset, which provides body motions and almost static hand motions. To fill in the face parameters, we extract facial expressions from the facial dataset BAUM~\cite{baum} with the SOTA facial reconstruction method EMOCA~\cite{emoca} and perform a data augmentation (in Sec.~\ref{sec:motion_aug}). For the text labels, we utilize the semantic labels from HumanML3D~\cite{humanml3d} and annotate the pose description with our whole-body pose captioning module.

\textbf{IDEA400} is a high-quality and expressive motion dataset recorded by ourselves, providing 13K motion sequences and 2.6M frames. Inheriting the NTU120~\cite{liu2019ntu} categories, we expand them to 400 actions with additional human self-contact motions, human-object contact motions, and expressive whole-body motions (e.g., rich facial expressions and fine-grained hand gestures). There are 36 actors with diverse appearance and clothing. For each action, we have the actors perform three times standing, three times walking, and four times sitting, a total of ten times. 
We annotate the SMPL-X format pseudo labels with the proposed motion annotation pipeline, which can generate high-quality whole-body motions. To obtain the semantic labels, we use the designed action labels and expand them with the large language model (LLM)~\cite{jiao2023chatgpt}. The pose descriptions are annotated with our whole-body pose captioning method. 
In this subset, we provide monocular videos, the body keypoints, SMPL-X parameters, and action labels. Please see the video in our website for more details and visualization. 

\textbf{AIST}~\cite{tsuchida2019aist} is a large-scale dance dataset with multi-view videos and dance genres labels. Instead of using the body-only SMPL annotations from AIST++~\cite{li2021aist}, we annotate the whole-body motion via our motion annotation pipeline. We obtain the semantic labels by expanding the dance genres label and providing frame-level pose descriptions.

\textbf{HAA500}~\cite{chung2021haa500} is a large-scale human-centric atomic action dataset with manually annotated labels and videos for action recognition. It contains 500 classes with fine-grained atomic action labels, covering sports, playing musical instruments, and daily actions. However, it does not have the motion labels. We annotate the 3D whole-body motion with our pipeline. We use the provided atomic action label as semantic labels. Besides, we input the video into video-LLaMA~\cite{damonlpsg2023videollama} and filter the human action descriptions as supplemental texts. Pose description is generated by our automatic pose annotation method.

\textbf{HuMMan}~\cite{cai2022humman} is a human dataset with multi-modality data, including multi-view videos, keypoints, SMPL parameters, action labels, etc. It does not provide whole-body pose labels. We estimate the SMPL-X parameters with our annotation pipeline. Besides, we expand the action label with LLM into semantic labels and use the proposed captioning pipeline to obtain pose descriptions.

\textbf{GRAB}~\cite{taheri2020grab} is human grasping dataset with body and hand motion. Meanwhile, it provides text descriptions of each grasping motion without corresponding videos. Therefore, similar to AMASS, we extract facial expressions from BAUM to fill in the facial expression. We use the provided text description as semantic labels and annotate pose descriptions based on the SMPL-X parameters.

\textbf{EgoBody}~\cite{zhang2022egobody} is a large-scale dataset capturing ground-truth 3D human motions in social interactions scenes. It provides high-quality body and hand motion annotations, lacking facial expression. Thus, we perform a motion augmentation to obtain expressive whole-body motions. Since EgoBody does not provide text information, we manually label the semantic description using the VGG Image Annotation (VIA)~\cite{dutta2019vgg} and annotate the pose description with the automatic pose captioning pipeline.

\textbf{BAUM}~\cite{baum} is a facial dataset with 1.4K audio-visual clips and 13 emotions. We annotate facial expressions from BAUM with the SOTA face reconstruction method EMOCA.

\textbf{Online Videos.} To improve the richness of appearance and motion diversity, especially on professional motions, we collect 33K monocular videos from online sources, covering various real-life scenes. We design action categories as motion prompts and input them into LLM. Then, we collect videos from online sources based on the answer of LLM, after which we filter the candidate videos by transition detection and annotate the whole-body motion, semantic label and pose description for the selected videos.

\subsection{Motion Augmentation Mechanism}
\label{sec:motion_aug}

\begin{figure*}[h]
    \begin{center}
        \includegraphics[width=1\textwidth]{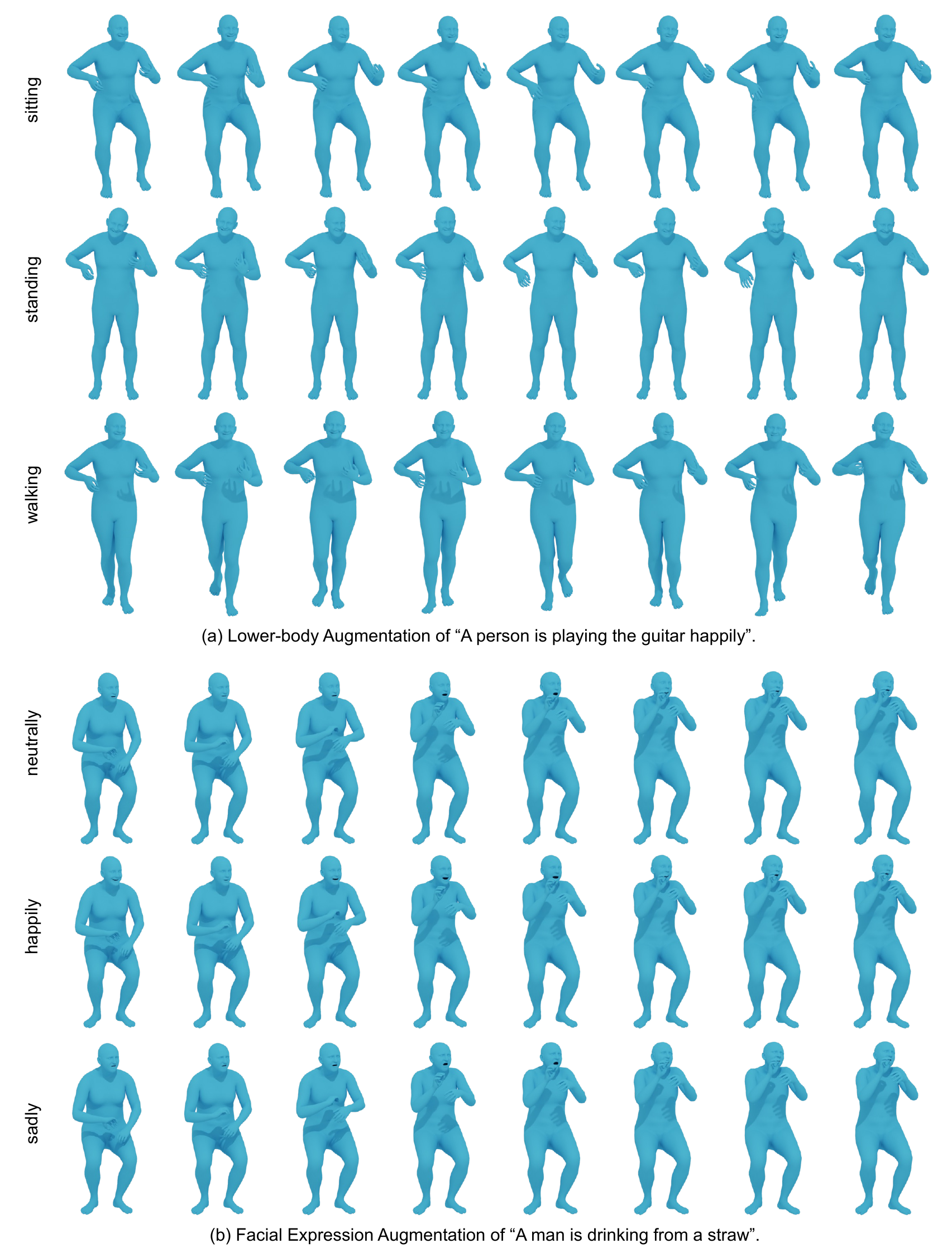}
    \end{center}
    \caption{Illustration of two motion augmentation methods: (a) lower-body augmentation. (b) facial expression augmentation.}
    \label{figure:motion_augmentation}
    \end{figure*}

\noindent\textbf{Lower-body Motion Augmentation.} \dataname contains some upper-body videos collected from online videos, like the videos in UBody~\cite{osx}, where the lower-body part is invisible. Estimating accurate lower-body motions and global trajectories for these videos is challenging. Thanks to the precise low-body motions provided in AMASS, we can simply perform a lower-body motion augmentation for these sequences, i.e., selecting the closest motion from AMASS based on the SMPL-X parameters and replacing the lower-body motion with it. Meanwhile, we incorporate relevant keywords (e.g., sitting, standing, walking) in the text descriptions. Fig.~\ref{figure:motion_augmentation}(a) depicts three plausible lower-body augmentations for the motion sequence with the semantic label "a person is playing the guitar happily." 

\noindent\textbf{Facial Expression Augmentation.} As shown in Tab.~\ref{tab:statistic_subdataset}, the motion capture datasets AMASS, GRAB, and EgoBody do not provide facial expressions. Thus, we perform a facial expression augmentation for these motions by randomly selecting a facial expression sequence from the BAUM~\cite{baum} dataset to fill the void and incorporating emotion labels (e.g., happy, sad, and surprise) in the semantic description. We perform interpolation for the selected sequence to ensure the same length as the original motion. An example of face expression augmentation is illustrated in Fig.~\ref{figure:motion_augmentation}(b).

\section{More Annotation Visual Results}
In this section, we present some visual results of the 2D keypoints, SMPL-X parameters, and motion sequences to show the effectiveness of our proposed motion annotation pipeline. 

\subsection{2D Keypoints}
\begin{figure*}[h]
    \begin{center}
        \includegraphics[width=1\textwidth]{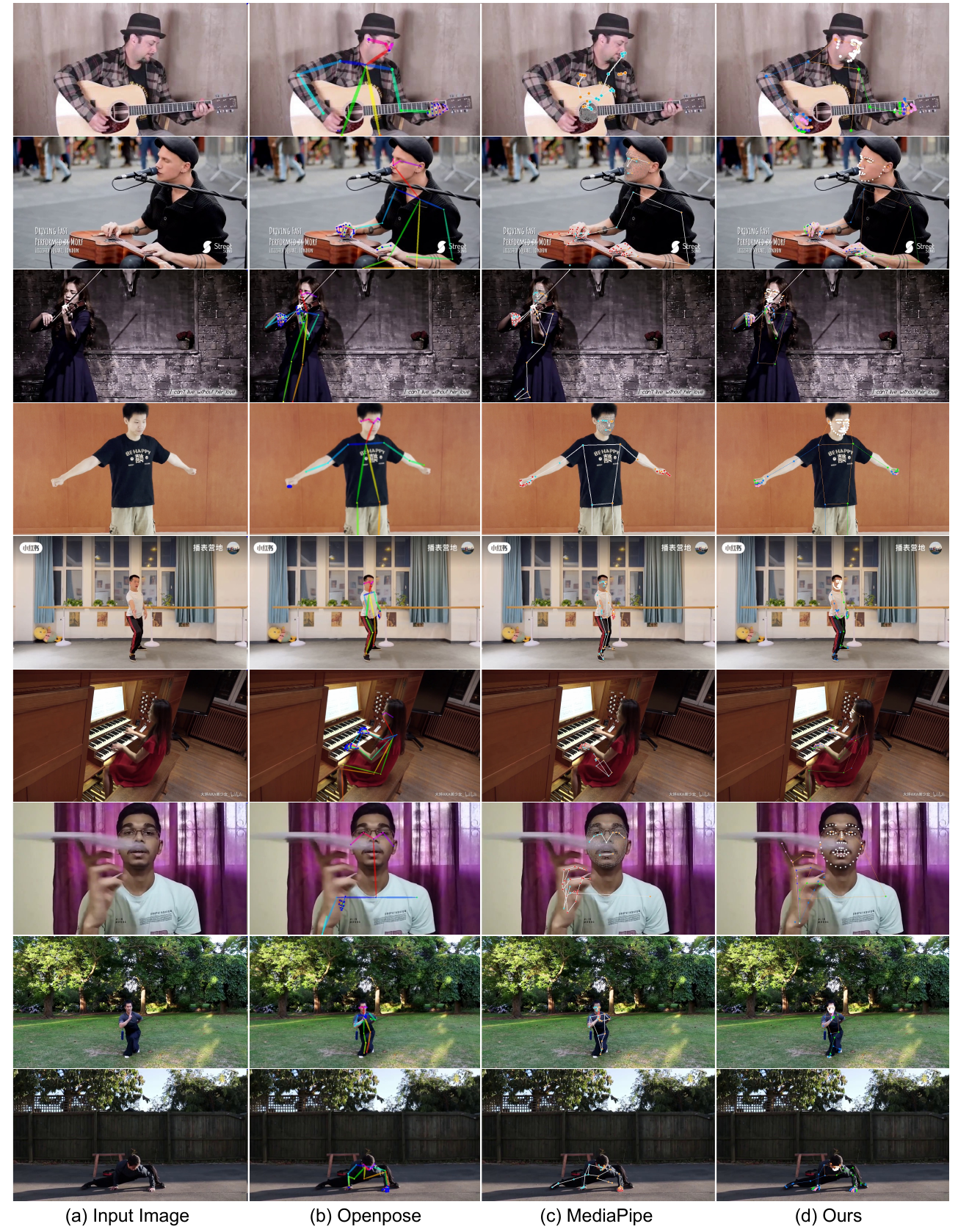}
    \end{center}
    \caption{Comparisons of the 2D keypoints annotation quality with widely used methods Openpose~\cite{openpose} and MediaPipe~\cite{mediapipe}.}
    \label{figure:kpt_annot_compare}
    \end{figure*}

As the main paper claims, we propose a hierarchical Transformer-based model for 2D keypoints estimation. To demonstrate the superiority of our method, we compare it with two  widely used methods, Openpose~\cite{openpose} and MediaPipe~\cite{mediapipe}. We use the PyTorch implementation of Openpose and only estimate the body and hand keypoints as it does not provide the face estimator. As shown in Fig.~\ref{figure:kpt_annot_compare}, Openpose and MediaPipe can not achieve accurate results in some challenging poses. Besides, there exists severe missing detection of hands for Openpose and MediaPipe. In contrast, our method performs significantly better, especially the hand keypoint localization.

\subsection{SMPL-X Parameters}
\begin{figure*}[h]
    \begin{center}
        \includegraphics[width=1\textwidth]{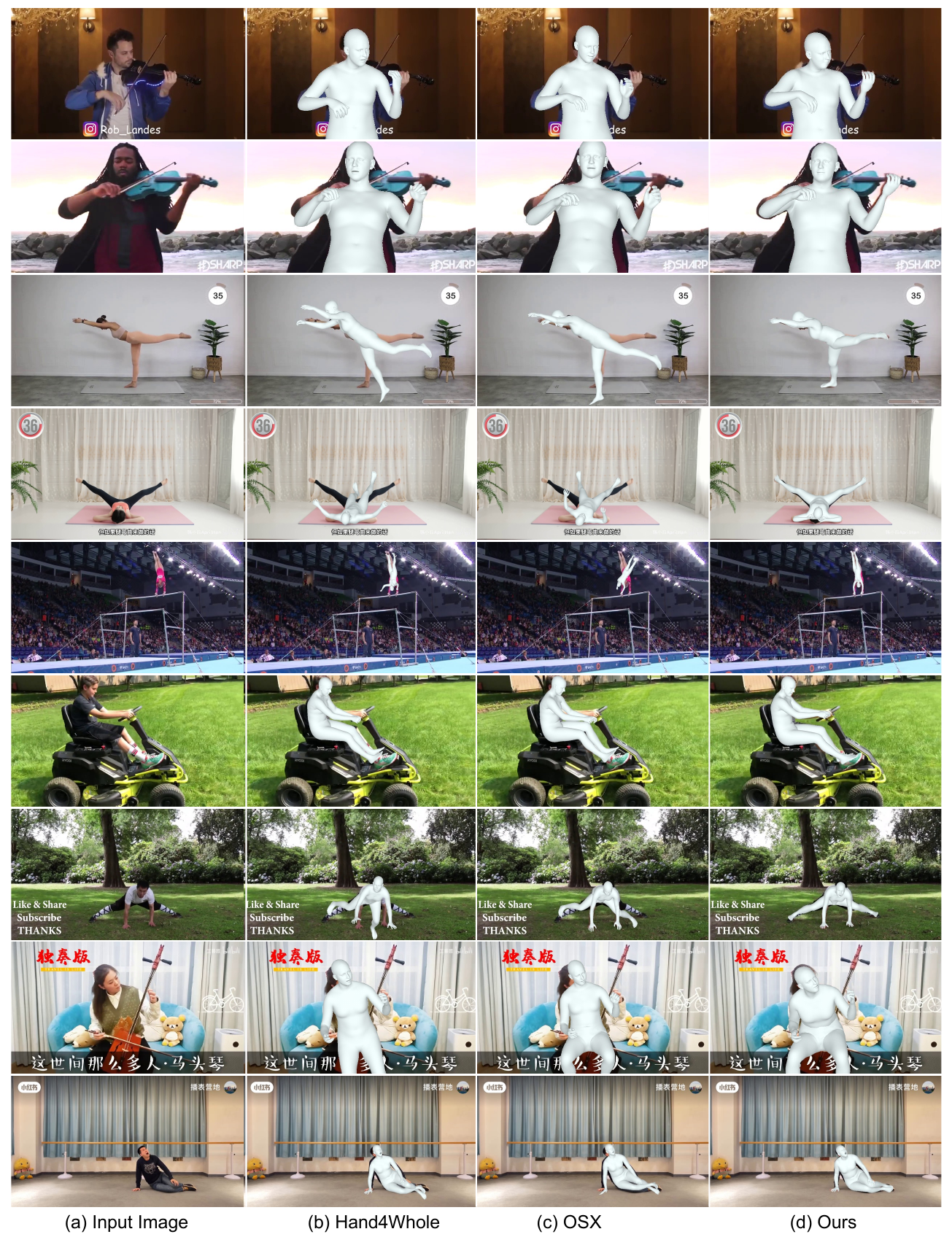}
    \end{center}
    \caption{Comparisons of the 3D SMPL-X annotation quality with SOTA methods Hand4Whole~\cite{GyeongsikMoon2020hand4whole} and OSX~\cite{osx}.}
    \label{figure:smplx_annot_compare}
    \end{figure*}

\begin{figure*}[h]
    \begin{center}
        \includegraphics[width=1\textwidth]{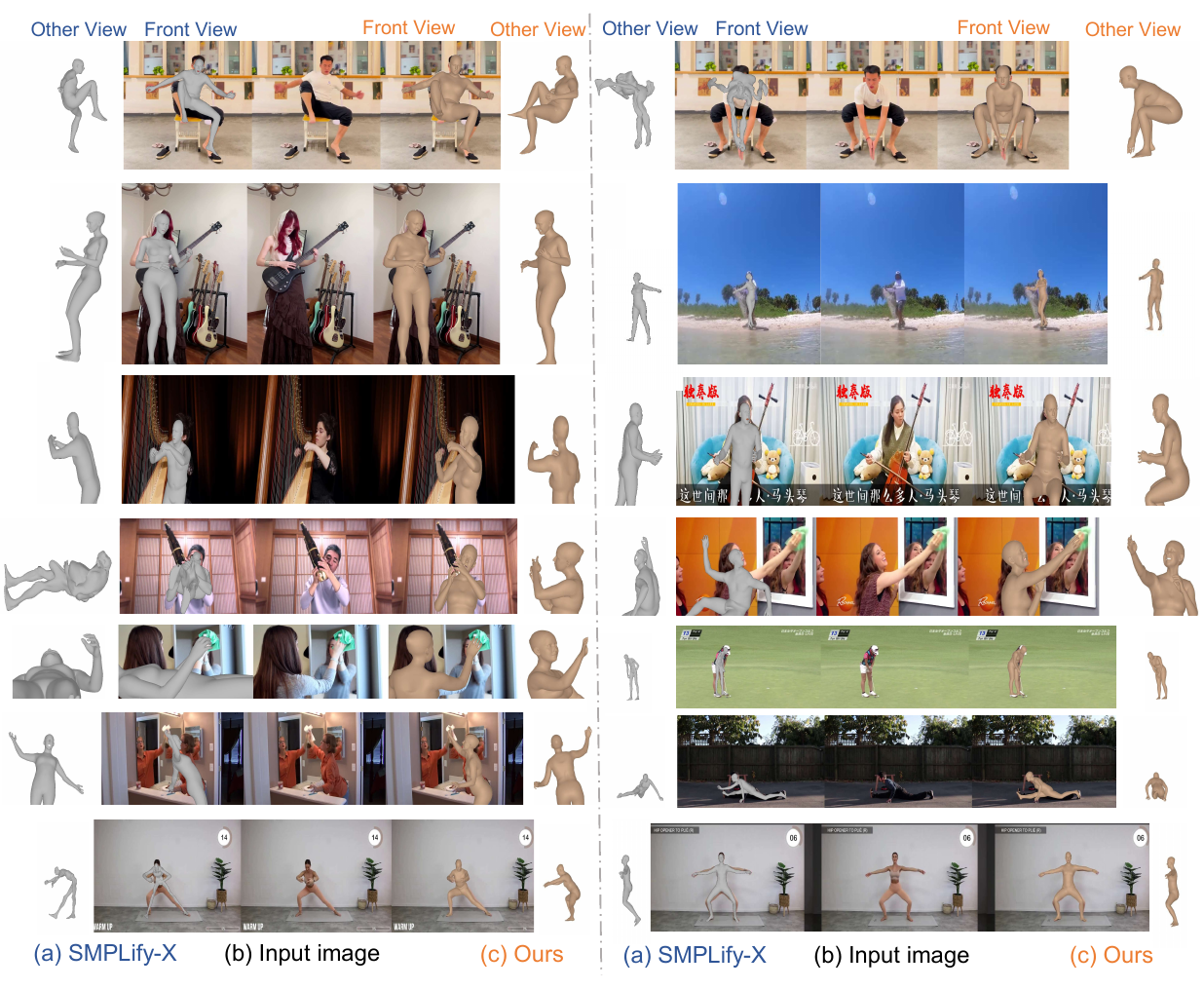}
    \end{center}
    \caption{Comparisons of the 3D SMPL-X annotation quality with wide-used fittiing methods SMPLify-X~\cite{smpl-x}.}
    \label{figure:comp_smplyx}
    \end{figure*}

To register accurate SMPL-X parameters, we elaborately design a learning-based fitting method with several training loss functions. We compare our method with two SOTA learning-based methods, Hand4Whole~\cite{GyeongsikMoon2020hand4whole} and OSX~\cite{osx}. As shown in Fig.~\ref{figure:smplx_annot_compare}, our method achieves a much better alignment result than the other models, especially on some difficult poses, which benefits from the iterative fitting process. Notably, Hand4Whole~\cite{GyeongsikMoon2020hand4whole} and OSX~\cite{osx} can only estimate the local positions without optimized global positions, which will suffer from unstable and jittery global estimation.
Furthermore, we compare with the widely used fitting method SMPLify-X~\cite{smpl-x}, using their officially released codes, in Fig.~\ref{figure:comp_smplyx}. Our method is more robust than SMPLify-X and can obtain better results about physically plausible poses, especially in challenging scenes (e.g., hard poses, low-resolution inputs, heavy occlusions). The results from the side view demonstrate that our method can properly deal with depth ambiguity and avoid the lean issue.

\subsection{Motion Sequences}
\begin{figure*}[h]
    \begin{center}
        \includegraphics[width=1\textwidth]{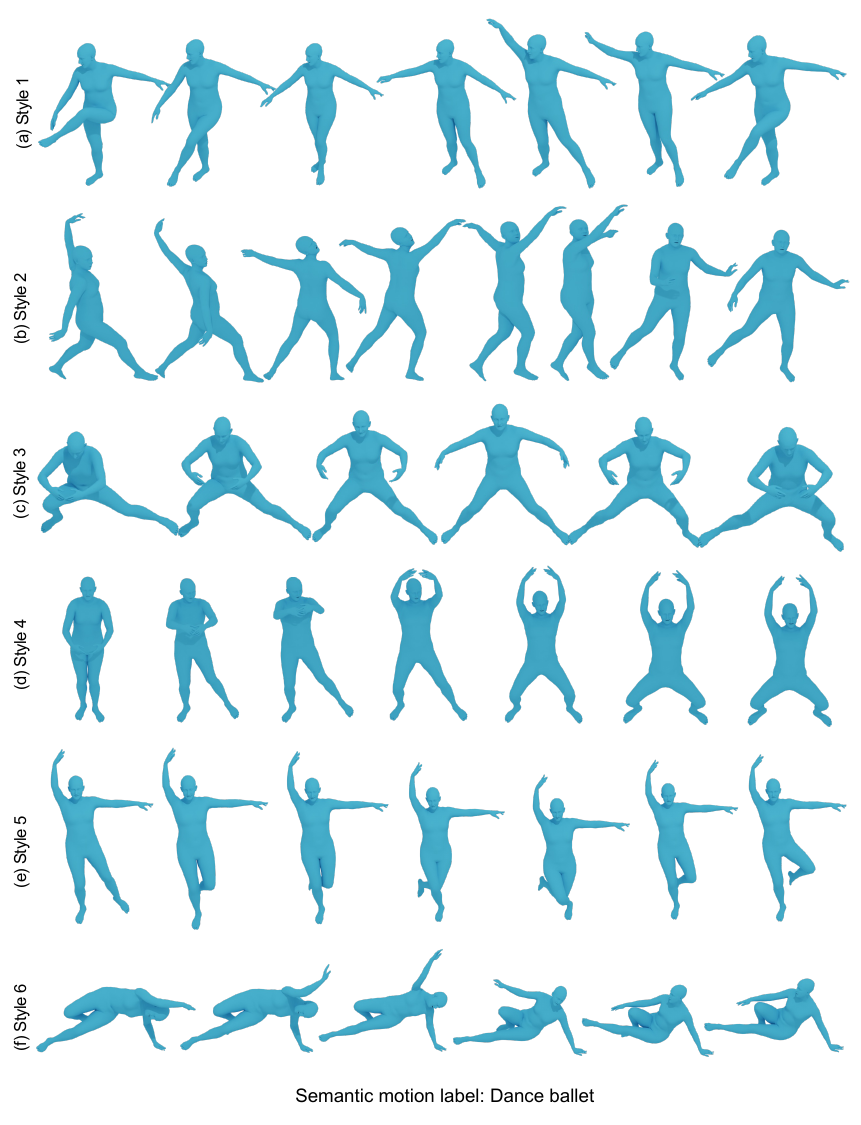}
    \end{center}
    \caption{Examples of the same semantic label ``dance ballet'' with different styles to enhance the motion diversity.}
    \label{figure:ballet}
    \end{figure*}

\begin{figure*}[h]
    \begin{center}
        \includegraphics[width=1\textwidth]{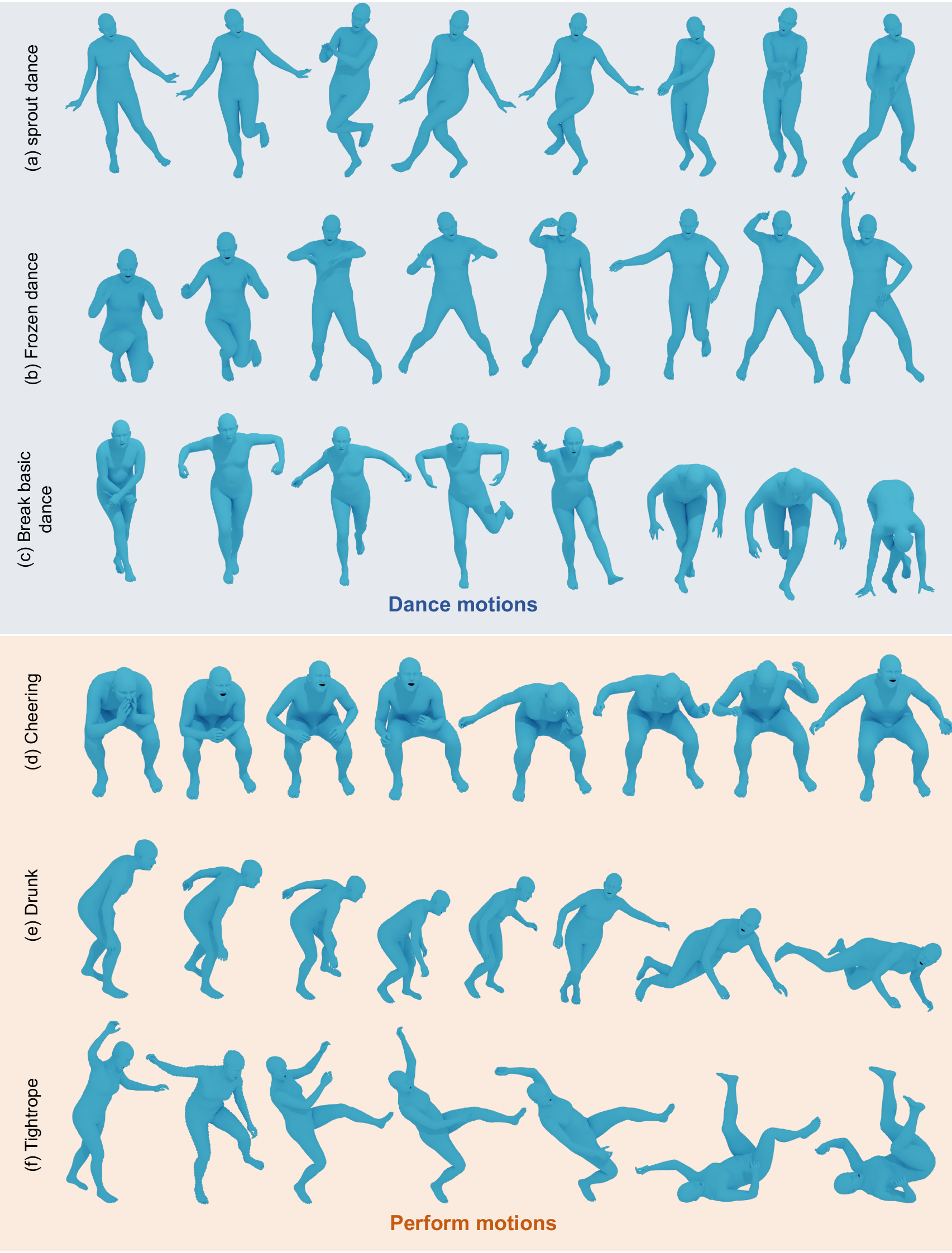}
    \end{center}
    \caption{Visualization of the dance and perform motion sequences of \dataname.}
    \label{figure:supp_viz1}
    \end{figure*}

\begin{figure*}[h]
    \begin{center}
        \includegraphics[width=1\textwidth]{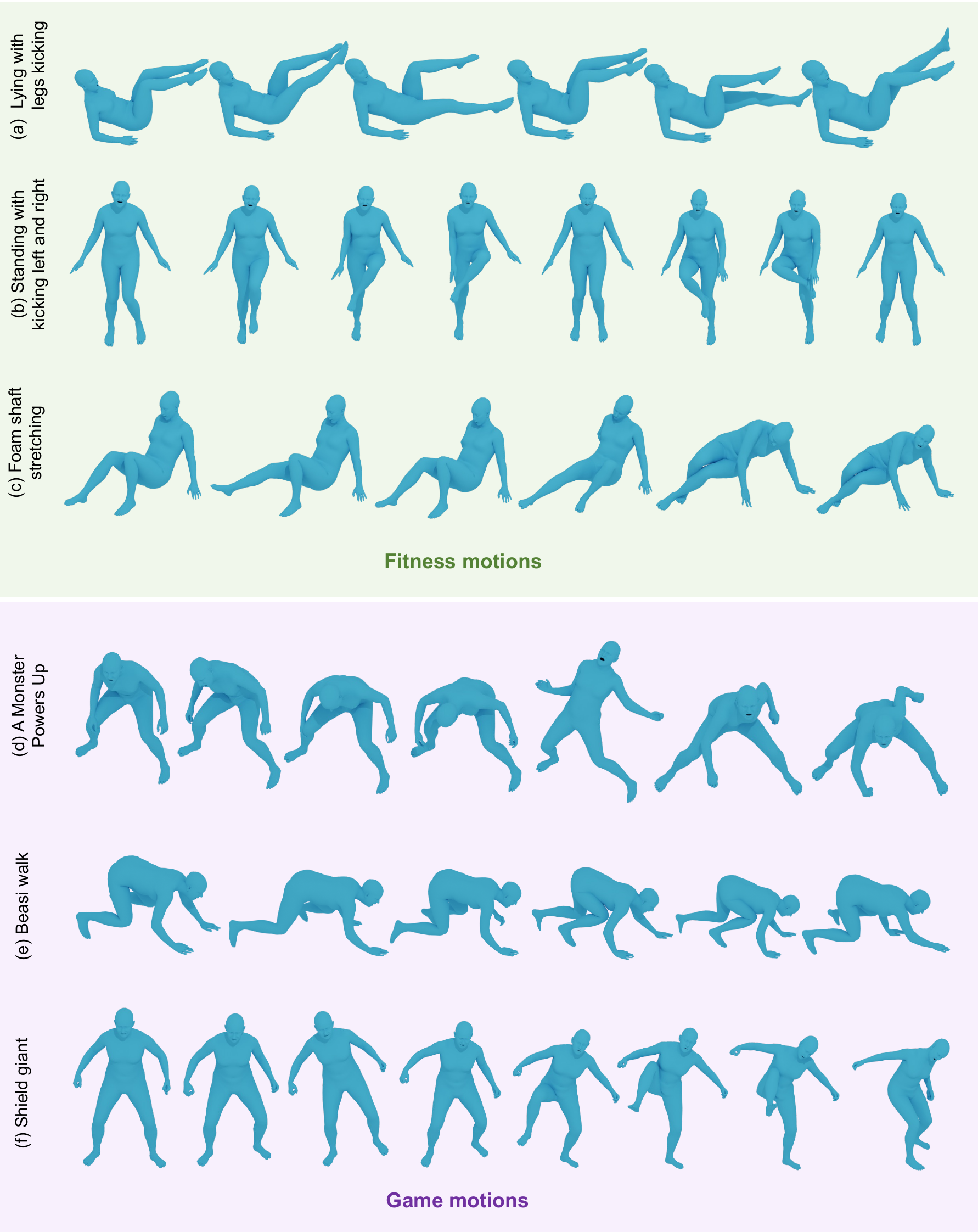}
    \end{center}
    \caption{Visualization of the fitness and game motion sequences of \dataname.}
    \label{figure:supp_viz2}
    \end{figure*}

To highlight the expressiveness and diversity of our proposed motions, we illustrate examples of the same semantic label, like \emph{dance ballet}, with six motion styles in Fig.~\ref{figure:ballet}. This one-to-many (text-to-motion) information can benefit the diversity of motion generation.
Then, we demonstrate more motion visualization in Fig.~\ref{figure:supp_viz1} and \ref{figure:supp_viz2} for different motion scenes. These motions show different facial expressions, hand poses, and body motions.

\section{Experiment}

\subsection{Experiment Setup}

\label{sec:exp_detail}
\noindent \textbf{Motion Representation.} To capture the 3D expressive whole-body motion, we use SMPL-X~\cite{smpl-x} as our motion representation. A pose state is formulated as:
\begin{equation}
    \mathbf{x} = \{\theta_b, \theta_h, \theta_f, \psi, \mathbf{r} \}.
\end{equation}
Here, $\theta_b\in\mathbb{R}^{22\times3}$ and $ \theta_h\in\mathbb{R}^{30\times3}$ denote the 3D body rotations and hand rotations. $\theta_f\in \mathbb{R}^{3}$ and $\psi\in \mathbb{R}^{50}$ are the jaw pose and facial expression. $\mathbf{r}\in\mathbb{R}^3$ is the global translation.

\noindent \textbf{Evaluation Metrics.} We adopt the same evaluation metrics as \cite{humanml3d,guo2020action2motion}, including Frechet Inception Distance (FID), multimodality, diversity, R-precision, and multimodal distance. We pretrain a motion feature extractor and a text feature extractor for the new motion presentation with contrastive loss to map the text and motion into feature space and then evaluate the distance between the text-motion pairs. For each generated motion, its ground-truth text description and 31 mismatched text description randomly selected from the test set compose a description pool. We rank the Euclidean distances between the generated motion and each text in the pool and then calculate the average accuracy at the top-k positions to derive R-precision. Multimodal distance is computed as the Euclidean distance between the feature vectors of generated motion and its corresponding text description in the test set. Additionally, We include the average temporal standard deviation as a supplementary metric to evaluate the diversity and temporal variation of whole-body motion.

\noindent \textbf{Computational Costs.} We use 8 NVIDIA A100 GPUs for motion annotation and 4 GPUs for motion generation experiments. It takes about 72 hours to annotate 1M frames with our annotation pipeline. 

\subsection{More Ablation Study}
\textbf{More Comparisons with HumanML3D.} 
Previous motion generation datasets are limited in expressing rich hand and face motions, as they only contain body and minimal hand movements. To demonstrate the expressiveness of our dataset, we conduct a comparison between HumanML3D and \dataname on face, hand, and body, separately. Specifically, we train MLD~\cite{mld} on each dataset and evaluate the diversity of generated motions and ground-truth motions by computing the average temporal standard deviation of the SMPL-X parameters and joint positions. 
The SMPL-X parameters include body poses, hand poses, and facial expressions. Body, hand, and face joint positions are represented as root-relative, wrist-relative, and neck-relative, respectively. We randomly choose 300 generated samples from the validation set and repeat the experiment 10 times to report the average results. As shown in Tab.~\ref{tab:ablation_motion}, the generated and ground-truth motions in \dataname exhibit a higher deviation, especially in hand and face parameters, indicating significant hand and face movements over time. These results demonstrate that the model trained with \dataname can generate more diverse facial expressions and hand motions, demonstrating the ability of our whole-body motion to capture fine-grained hand and face movements and expressive actions.

\begin{table*}[h]
	\vspace{4pt}
        \centering
	\resizebox{0.6\textwidth}{!}{%
    \begin{tabular}{l|ccc|ccc}
        \toprule
        \multicolumn{1}{c|}{\multirow{2}[4]{*}{{Method}}} & \multicolumn{3}{c|}{\boldmath{}{Joints Position $\uparrow$}\unboldmath{}} & \multicolumn{3}{c}{\boldmath{}{SMPL-X Param ($10^{-2}$) $\uparrow$}\unboldmath{}} \\
        \cmidrule{2-7}
        & {Face} & {Hand} & {Body} & {Face} & {Hand} & {Body} \\
        \midrule
        HumanML3D (GT) & 0.00 & 0.00 & 88.7 & 0.00 & 0.00 & 9.01  \\
        Motion-X (GT) & 1.60 & 8.82 & 92.2 & 13.4 & 5.24 & 9.95  \\
        \midrule
        HumanML3D &0.00 & 0.00 & 64.8 & 0.00 & 0.00 & 7.23  \\
        Motion-X & 1.33 & 11.4 & 66.0 & 7.28 & 5.30 & 7.41  \\
        \bottomrule
        \end{tabular}
	}
    \caption{\small Temporal standard deviation of the SMPL-X parameters and joint positions on HumanML3D and \dataname. We compare the GT and generated motions with the MLD model trained on HumanML3D and \dataname, respectively.}	       
    \label{tab:ablation_motion}
\vspace{-0.3cm}
\end{table*}

\section{Related Work}

In this part, we introduce relevant \textbf{methods} for human motion generation.

According to different inputs, producing human motions can be divided into two categories: the general motion synthesis from scratch~\cite{yan2019convolutional,zhao2020bayesian,zhang2020perpetual,cai2021unified} and the controllable motion generation from given text, audio, and music as conditions~\cite{ahn2018text2action,temos,modiff,motiondiffuse,mld,humanml3d,ahuja2019language2pose,ghosh2021synthesis}. 
Motion synthesis encompasses several tasks, such as motion prediction, completion, and interpolation~\cite{cai2021unified}, developed over several decades in computer vision and graphics. These tasks tend to utilize nearby frames with spatio-temporal correlations to infer estimated frames in a deterministic manner~\cite{martinez2017human,butepage2017deep,cai2020learning,fragkiadaki2015recurrent,ghosh2017learning,kaufmann2020convolutional,mao2019learning,mao2020history}. 
On the other hand, motion generation is a more challenging task that aims to synthesize long-term, diverse, natural human motions. 

Many generative models, like GANs, VAEs, and recent diffusion models, have been explored~\cite{wang2020learning,yu2020structure,action2video,guo2020action2motion,petrovich2021action}.
This work mainly discusses text-conditioned motion generation. This field has evolved from inputting action classes~\cite{action2video,guo2020action2motion} to sentence descriptions~\cite{humanml3d,mld,motiondiffuse}, and generating motions from 2D to 3D keypoints, to the emerging parametric model (e.g., SMPL~\cite{smpl,humanml3d,mld}). These models have become expressive and comprehensive toward real-world scenarios thanks to the development of related benchmarks. 
Recently, diffusion model-based methods have rapidly developed and shown advantages in diverse, realistic, and fine-grained motion generation~\cite{motiondiffuse,mld,modiff,Tevet2022HumanMD,Yuan2022PhysDiffPH}. Some concurrent works~\cite{Tevet2022HumanMD,motiondiffuse,mld} introduce novel diffusion model-based motion generation framework to achieve state-of-the-art (SOTA) quality. For example, MLD~\cite{mld} presents a motion latent-based diffusion model with a representative motion variational autoencoder, showing its efficiency. Based on the proposed Motion-X, HumanTOMATO~\cite{humantomato} introduces the first text-aligned whole-body motion generation that can generate high-quality, diverse, and coherent facial expressions, hand gestures, and body motions simultaneously.

\section{Limitation and Broader Impact}
\subsection{Limitation} There are two main limitations. \textbf{(i)} The motion quality of our markless motion annotation pipeline is inevitably inferior to the multi-view mark-based motion capture system. However, as the quantitative and qualitative results demonstrate, our method can perform much better than existing markless methods, thanks to large-scale models pre-trained on massive 2D and 3D keypoints datasets and our elaborately designed fitting pipeline. Besides, a 30 mm PA-MPVPE error would be acceptable for the text-driven motion generation task since the target is to synthesize natural and realistic motions that are semantically consistent with the text input. Furthermore,
the experiment on the mesh recovery task has demonstrated that our dataset can also benefit the human reconstruction task, which requires a higher annotation quality.
Accordingly, a better motion annotation will be beneficial, and we will leave it as our future work.
\textbf{(ii)} During our experiment, we found out that existing evaluation metrics are not always consistent with visual results. Besides, SMPL-X parameters may not be the best motion representation for expressive whole-body motion representation. Thus, there is a need for further research on the evaluation metric, motion representation, and model designs for the expressive motion generation task. We leave them as future work.

\subsection{Broader Impact}

Based on the scalable and automatic annotation way proposed in this work, although there are inevitable errors, large-scale data could be helpful. Meanwhile, this way can boost the direction of ``learning from noisy labels" for related tasks, such as text-driven whole-body motion generation. 
A large-scale 3D human motion dataset would have numerous applications and boost novel research topics in various fields, such as animation, games, virtual reality, and human-computer interaction. Until now, human motion datasets have had no negative social impact yet. Our proposed \dataname will strictly follow the license of previous datasets, and would not present any negative foreseeable societal consequence, either.

\section{License}
All data is distributed under the CC BY-NC-SA (Attribution-NonCommercial-ShareAlike) license. Detailed license and instructions can be found on the page \url{https://motion-x-dataset.github.io}.  Further, we will provide a GitHub repository to solicit possible annotation errors from data users.
For the sub-datasets, we would ask the user to read the original license of each original dataset, and we would only provide our annotated result to the user with the approvals from the original Institution. Here, we provide a brief license of the used assets:

\begin{itemize}
    \item HumanML3D dataset~\cite{humanml3d} originates from the HumanAct12~\cite{guo2020action2motion} and AMASS~\cite{amass} datasets, which are both released for academic research only and it is free to researchers from educational or research institutes for non-commercial purposes. 
    \item BAUM dataset~\cite{baum} is CC-BY 4.0 licensed.
    \item HAA500 dataset~\cite{chung2021haa500} is MIT licensed. 
    \item IDEA400 dataset belongs to the International Digital Economy Academy (IDEA) and is licensed under the Attribution-Non Commercial-Share Alike 4.0 International License (CC-BY-NC-SA 4.0).
    \item HuMMan dataset~\cite{cai2022humman}  is under S-Lab License v1.0.
    \item AIST dataset~\cite{li2021aist} is CC-BY 4.0 licensed.
    \item GRAB dataset~\cite{taheri2020grab} is released for academic research only and is free to researchers from educational or research institutes for non-commercial purposes.
    \item EgoBody~\cite{zhang2022egobody} is under CC-BY-NC-SA 4.0 license. 
    \item Other data is under CC BY-SA 4.0 license.
    \item SMPLify-X~\cite{smpl-x} codes are released for academic research only and are free to researchers from educational or research institutes for non-commercial purposes.
    \item Codes for preprocessing and training are under MIT LICENSE.

\end{itemize}

\clearpage

{
\bibliographystyle{ieeetr}
\bibliography{reference}
}

\end{document}